\documentclass[journal]{IEEEtran}
\ifCLASSINFOpdf
   \usepackage[pdftex]{graphicx}
\else
\fi
\usepackage[font=small]{caption}
\usepackage{amsmath}
\usepackage{amssymb}
\usepackage{amsthm}
\newtheorem{remark}{Remark}
\newtheorem{definition}{Definition}
\newtheorem{example}{Example}
\DeclareMathOperator{\diag}{diag}
\DeclareMathOperator{\Tr}{Tr}
\usepackage{subcaption} 
\usepackage{booktabs}
\usepackage[ruled,vlined]{algorithm2e}
\usepackage{algpseudocode}
\usepackage{scalerel}
\usepackage{verbatim}
\usepackage{commath}
\SetCommentSty{mycommfont}
\DeclareMathOperator*{\bigcircledast}{\scalerel*{\circledast}{\bigodot}}
\algdef{SE}{Begin}{End}{\textbf{begin}}{\textbf{end}}
\newcommand{\sizecorr}[1]{\makebox[0cm]{\phantom{$\displaystyle #1$}}}

\begin{document}
%
\title{Supervised Learning for Non-Sequential Data: A Canonical Polyadic Decomposition Approach}
%
%
%

\author{Alexandros~Haliassos\textsuperscript{*},
        Kriton~Konstantinidis\textsuperscript{*},
        and~Danilo~P.~Mandic,~\IEEEmembership{Fellow,~IEEE}
\thanks{*Equal Contribution}
\thanks{All authors are with the Department of Electrical and Electronic Engineering, Imperial College London, London SW7 2AZ, U.K. (e-mails: \{alexandros.haliassos14, k.konstantinidis19, d.mandic\}@imperial.ac.uk) }}

\maketitle

\begin{abstract}

Efficient modelling of feature interactions underpins supervised learning for non-sequential tasks, characterized by a lack of inherent ordering of features (variables). The brute force approach of learning a parameter for each interaction of every order comes at an exponential computational and memory cost (Curse of Dimensionality). To alleviate this issue, it has been proposed to implicitly represent the model parameters as a tensor, the order of which is equal to the number of features; for efficiency, it can be further factorized into a compact Tensor Train (TT) format. However, both TT and other Tensor Networks (TNs), such as Tensor Ring and Hierarchical Tucker, are sensitive to the ordering of their indices (and hence to the features). To establish the desired invariance to feature ordering, we propose to represent the weight tensor through the Canonical Polyadic (CP) Decomposition (CPD), and introduce the associated inference and learning algorithms, including suitable regularization and initialization schemes. It is demonstrated that the proposed CP-based predictor significantly outperforms other TN-based predictors on sparse data while exhibiting comparable performance on dense non-sequential tasks. Furthermore, for enhanced expressiveness, we generalize the framework to allow feature mapping to arbitrarily high-dimensional feature vectors. In conjunction with feature vector normalization, this is shown to yield dramatic improvements in performance for dense non-sequential tasks, matching models such as fully-connected neural networks.

\end{abstract}



%
\IEEEpeerreviewmaketitle

\section{Introduction}
\IEEEPARstart{M}ODERN data sources often exhibit non-sequential forms, whereby the features (variables) do not possess an associated inherent ordering. For example, features for estimating house prices may be the number of bedrooms and bathrooms, location, garage capacity, to mention but a few. Other non-sequential tasks include fraud detection, credit assignment, and movie recommendation. Imposing a structure, such as locality, on the features of such tasks would usually result in physically meaningless and practically ill-conceived problem formulations. Notice a stark contrast with e.g., image recognition, where the spatial arrangement of the features (pixels) can be meaningfully exploited by learning systems, such as Convolutional Neural Networks (CNNs) \cite{krizhevsky2012imagenet}. 

Two of the standard models for regression or classification paradigms with non-sequential real-valued variables are Support Vector Machines (SVMs) and fully-connected Neural Networks (NNs); these do not assume any a priori feature ordering, while interactions between the variables are modelled via polynomial or radial basis function kernels for the former and hierarchical network structures for the latter. However, both SVMs and NNs are known to underperform for very sparse data (i.e., categorical data that are one-hot-encoded), which limits their application as out-of-the-box, general predictors \cite{rendle2010factorization}.

As a remedy, general predictors that can efficiently model interactions of both dense (real-valued) and sparse (categorical) features include the tensor-based work in \cite{rendle2010factorization, blondel2016higher, NIPS2016_6211, novikov2018exponential}. They model interactions between variables by mapping each of the $N$ features to a $d$-dimensional vector and subsequently \textit{implicitly} taking the outer product of the $N$ vectors to construct a high-order tensor. Such a multi-dimensional array contains all possible interactions between the features and the final prediction is performed (again implicitly) through an inner product of the so constructed rank-1 tensor with the corresponding weight tensor. For affordable computations and better generalization, the exponentially-scaled weight tensor is represented by a compact Tensor Network, i.e., a set of interconnected lower-order tensors.

Overall, the advantages of this framework include:
\begin{itemize}
    \item Suitability for both dense and sparse data;
    \item Linear scaling with the dimensionality of the feature vectors and constant scaling with the training size (assuming mini-batch gradient descent in training), thus alleviating the Curse of Dimensionality and making it suitable for Big Data applications;
    \item Enhanced interpretability, due to the multi-linear nature of the model; this is not possible to achieve with other state-of-the-art approaches such as Neural Networks and Support Vector Machines (SVMs);
    \item Inherent universal function approximation property for large enough dimension of the local feature maps;
    \item Potential to derive novel computationally tractable machine learning models using multilinear algebra.
\end{itemize}

The focus of our work is on a class of predictors first introduced in \cite{NIPS2016_6211, novikov2018exponential}. 
Unlike existing works, which use the Tensor Train (TT) format \cite{oseledets2011tensor} to represent the weight tensor, our analysis has shown that, for non-sequential data, the well-established Canonical Polyadic (CP) format \cite{carroll1970analysis, harshman1970foundations} is a perfect match for this paradigm. This choice is also motivated by the observation that the cores (lower-order tensors) of the TT format are ordered in a sequence (hence the name Tensor \textit{Train}), and permuting the order of the cores would change the underlying tensor. This implies that strict ordering must be imposed a priori on the inherently unordered feature variables (akin to imposing ordering when applying a Convolutional Neural Network or Recurrent Neural Network \cite{graves2013speech} to non-sequential data). Other popular TNs, such as Hierarchical Tucker \cite{hackbusch2009new} (a binary balanced Tensor Tree) and Tensor Ring \cite{zhao2016tensor} (a generalization of TT), also exhibit an inherent ordering, even though the latter is invariant to circular dimensional permutation. On the other hand, as desired, the permutation of factor matrices within the CP format (assuming unities on the superdiagonal) does not affect the representation (see Fig. \ref{fig:cp_and_tt}).

The model proposed in this work is referred to as the \textit{CP-based predictor} while the models based on TT are referred to as \textit{TT-based}. Overall, compared with the existing TT-based predictors, the use of CP format offers the following advantages for non-sequential data:

\begin{itemize}
    \item Robustness to feature permutations, to match non-sequential data processing;
    \item It allows for simpler optimization algorithms;
    \item More parsimonious and interpretable representations.
\end{itemize}

The contributions of this work are as follows. We analytically derive prediction and learning algorithms, as well as the corresponding initialization and regularization schemes, that scale linearly with the dimensionality of the features and the local dimension, a key property for computational efficiency. The proposed model is shown to significantly outperform other Tensor Network-based counterparts on challenging prediction from sparse data, namely the MovieLens 100K recommender system dataset.

In addition, previous work on this class of predictors employs a local mapping of the features to 2-dimensional vectors. (We refer to the dimension of these vectors as the \textit{local dimension}.) On the other hand, the ability to increase the local dimension plays a critical role in the expressiveness of the models (they can only enjoy the universal function approximation property if the local dimension is large enough) \cite{cohen2016expressive}. A higher local dimension within the existing framework would lead to significant computational bottleneck for learning algorithms that, unlike ours, do not scale linearly with the local dimension, such as in the work of \cite{NIPS2016_6211}. Higher local dimension would also lead to instability or poor generalization, depending on the feature map used, as is shown in Section \ref{sec:experiments}.

A further contribution of this work, that is agnostic to the type of Tensor Network format used, is therefore the generalization from the currently used 2-dimensional feature maps to arbitrary $d$-dimensional maps; in this way, the proposed framework is equipped with the ability to model all interactions of features raised to powers of up to $(d-1)$. We also introduce unit normalization of the feature vectors after local mapping and demonstrate through experiments that, by virtue of this normalization, a higher $d$ can dramatically enhance performance, while learning algorithms remain stable even for for a local dimension as high as $d>100$. The performance enhancement achieved through an increase in $d$ is shown to enable this class of models to exhibit competitive results over other popular models on \textit{dense} data, including SVMs and fully-connected neural networks.

The TN-based predictors have been implemented in \texttt{TensorFlow 2.0} to enable straightforward experimentation with various optimizers, regularizers, and loss functions.\footnote{https://github.com/KritonKonstantinidis/CPD\_Supervised\_Learning}

The rest of the paper is organized as follows. We first discuss related works in Section \ref{sec:related_works} and then present the tensor preliminaries necessary to follow this work in Section \ref{sec:prelim}. After introducing our proposed model in Section \ref{sec:cp_based_mod}, we present analytical derivations and algorithms for efficient model prediction and learning in Section \ref{Algorithm}. Local feature maps for different settings are subsequently considered in Section \ref{sec:feat_mappings} and procedures for order regularization \cite{novikov2018exponential} are derived in Section \ref{Regularization}. Model initialization is addressed in Section \ref{sec:linear_model_init}. Finally, we provide experimental results in Section \ref{sec:experiments}, and conclude with future research directions in Section \ref{sec:conclusion}.

\section{Related Works} \label{sec:related_works}
Related works can be categorized into: 1) research on applying tensor decompositions (TDs) and Tensor Networks (TNs) to machine learning; 2) other closely related general predictors for non-sequential data that are suitable for both dense and sparse data; 3) existing Tensor Network-based methods within our considered framework. For a more in-depth comparison with the most closely-related models, see Appendix \ref{app:related_works}.
\subsection{Tensors for Machine Learning}
\subsubsection{General Applications}
Given the inherent multi-way structure of many data types, TDs and TNs have been extensively studied in the context of machine learning \cite{sidiropoulos2017tensor}. For example, TDs can discover underlying structures in multi-way data, separate signals in blind source separation applications, disentangle factors of variations in facial images, and compress data using low-rank approximations, while also potentially increasing their signal-to-noise ratio \cite{acar2008unsupervised, cichocki2009nonnegative, vasilescu2002multilinear, kroonenberg2008applied, comon2009tensor}. 

Another application is feature extraction from data represented as multi-dimensional arrays, such as color images, videos, or fMRI data \cite{lu2008mpca, zhou2015efficient, li2016mr, shi2015semi}. More recently, in \cite{shi2018feature} features from tensors with missing entries are extracted using low-rank tensor decompositions and feature variance maximization. In contrast, our data comes in the form of standard feature vectors as in the typical supervised learning context, rather than the somewhat restricted case of multi-dimensional arrays. As such, in our framework a high-order tensor is not actually decomposed, but instead it is only \textit{implicitly assumed} that the high-order feature interactions can be represented in a CP format. Furthermore, we are concerned with regression or classification from these feature vectors, and not with extracting new features.

A related area of research is tensor-on-tensor regression, where an output tensor is predicted from an input tensor. The CP tensor regression model \cite{zhou2013tensor} aims to predict a vector given a multi-dimensional array as input; the corresponding weights are assumed to be in CP format and an alternating minimization scheme is employed. In \cite{li2018tucker} and \cite{hou2015hierarchical}, a Tucker and a Hierarchical Tucker format are respectively assumed. Very recently, generalized tensor-on-tensor regression is performed using the Tensor Train format \cite{liu2020low}. The key difference between these methods and our model lies again in the type of task considered. Whereas in \cite{zhou2013tensor, li2018tucker, hou2015hierarchical, liu2020low} the input data is a multi-dimensional array, our samples are feature vectors. Another key difference is that, unlike in tensor-on-tensor regression problems, our implicit tensor (which holds feature interactions) is rank-1 by construction, which, when coupled with the weight tensor in the CP format, leads to very efficient learning algorithms, as described in Section \ref{Algorithm}. 

\subsubsection{Tensors for Deep Learning}
Recent efforts in the deep learning literature propose to heavily compress fully-connected, convolutional, and recurrent networks, without a significant loss in performance \cite{lebedev2014speeding, calvi2019tucker,novikov2015tensorizing, wang2018wide}. In particular, the authors in \cite{lebedev2014speeding} decompose the 4-D kernels of Convolutional Neural Networks using CPD. This differs significantly from our approach, where CPD is not simply used to compress other models, but instead plays a central role in the model itself; that is, we represent an exponentially large tensor of rich feature interactions with the CPD. While the expressiveness of Neural Networks is due to their hierarchical structure, the expressiveness of our model is due to the implicit construction of this tensor. 

Tensor Networks have also been employed for the analysis of Neural Networks. By establishing links between common TNs, such as the Tensor Train (TT) \cite{oseledets2011tensor} or Hierarchical Tucker (HT) networks \cite{hackbusch2009new}, and deep learning architectures such as Recurrent \cite{graves2013speech, mandic2001recurrent} or Convolutional Neural Networks, it has been possible to obtain new theoretical insights related to the expressiveness of deep networks compared to their shallow counterparts \cite{cohen2016expressive,khrulkov2018expressive,khrulkov2019generalized}.

\subsection{Other General Predictors for Non-Sequential Data}
Support Vector Machines are a standard class of models for non-sequential data. They employ the well-known kernel trick and are optimized in their dual form. They also require storing training data points (support vectors) for inference. More importantly, they tend to underperform when dealing with very sparse data \cite{rendle2010factorization}. 

Factorization Machines (FMs) and their extension, Higher-Order FMs (HOFMs), address these problems by modelling interactions using factorized parameters \cite{rendle2010factorization, blondel2016higher}, whereby FMs can model pairwise interactions while HOFMs can model interactions of higher orders. Polynomial Networks additionally include interactions of features raised to powers larger than one \cite{livni2014computational}. In contrast, our predictor captures all interactions of \textit{every} order in linear time during training and inference, and it is able to handle interactions of any \textit{arbitrary function} of features.

\subsection{Other Works Under the Considered Framework}
The framework upon which our work is based was first introduced in \cite{ NIPS2016_6211, novikov2018exponential}, where the weight tensor is represented as a Tensor Train, and the features are mapped to 2-dimensional vectors. In \cite{NIPS2016_6211}, a trigonometric basis is used for feature mapping, while the model is trained using a sweeping algorithm, inspired by DMRG \cite{schollwock2011density}; this makes it difficult to apply it in the stochastic gradient setting as it scales cubically with the local dimension. The authors of \cite{novikov2018exponential} designed a stochastic version of a Riemannian optimization approach, which they found to be more robust to initialization than stochastic gradient descent methods. A similar recent method uses a MERA-inspired algorithm \cite{cichocki2017tensor} and draws comparisons with quantum physics \cite{Liu_2019}; it is shown that MERA is suitable for image recognition tasks due to its 2-D structure, whereas in our work we consider non-sequential tasks.

Our work differs in several substantial ways. First, motivated by the importance of invariance to the ordering of the features, we propose to represent the weight tensor in the CP format for non-sequential tasks, rather than the TT format in \cite{novikov2018exponential}, which requires rigid feature ordering; this intuition is validated through experiments on both dense and sparse data. We have also implemented, for the first time in this framework, the corresponding models based on the Tucker, Tensor Ring, and Hierarchical Tucker formats, with our proposed model consistently outperforming these newly considered models in all experiments.
In addition, to provide deeper insights and at the same time establish a platform for further research, we have analytically derived, using standard tensor algebra notation, the learning algorithms and regularization schemes associated with our proposed predictor in the stochastic gradient descent setting. We also prove how our CP-based predictor can be initialized to correspond to the linear model solution, an important aspect for the success of the model, as demonstrated on the MovieLens dataset. We further study the effects of exceeding the restricted setting of feature maps of local dimension of 2, and show that \textit{as long as the mapped vectors are normalized}, the performance of these predictors can be dramatically increased if we vary the local dimension. This change has enabled us to obtain competitive performance with fully-connected neural networks on a task with dense data, which was not possible in existing works, where the local dimension was fixed to 2 (see Fig. \ref{fig:high_d}). 

\section{Notations and Preliminaries} \label{sec:prelim}

\subsection{Tensor Notation and Basic Operations}
A real-valued tensor is a multidimensional array, denoted by a calligraphic font, e.g., $\mathcal{X}\in\mathbb{R}^{I_1\times\dots\times I_N}$, where $N$ is the order of the tensor, and $I_n$ ($1 \leq n \leq N$) the size of its $n$\textsuperscript{th} mode. Matrices (denoted by bold capital letters, e.g., $\mathbf{X}\in\mathbb{R}^{I_1\times I_2}$) can be seen as second order tensors ($N=2$), vectors are denoted by bold lower-case letters, e.g., $\mathbf{x}\in\mathbb{R}^{I}$ and can be seen as order-1 tensors ($N=1$), and scalars (denoted by lower-case letters, e.g., $x\in\mathbb{R}$) are tensors of order $N=0$. A specific entry of a tensor $\mathcal{X}\in\mathbb{R}^{I_1\times\dots\times I_N}$ is given by $x_{i_1,\dots,i_N}\in\mathbb{R}$. Moreover, we adopt a graphical notation, whereby a tensor is represented by a shape (e.g., a circle) with outgoing edges; the number of edges equals the order of the tensor (see \cite{tensornetworksdimens} for more information on this graphical notation).

The following conventions for basic linear/multilinear operations are employed throughout the paper. 
\begin{definition}[Multi-Index] A multi-index (in reverse lexicographic ordering) is defined as $\overline{i_1i_2\dots i_N}=i_1+(i_2-1)I_1+(i_3-1)I_1I_2+\dots+(i_N-1)I_1\dots I_{N-1}$.
\end{definition}
\begin{definition}[Tensor Matricization] The mode-$n$ matricization of a tensor $\mathcal{X}\in\mathbb{R}^{I_1\times\dots\times I_N}$ reshapes the multidimensional array into a matrix $\mathbf{X}_{(n)}\in\mathbb{R}^{I_n\times I_1I_2\dots I_{n-1}I_{n+1}\dots I_N}$ with $(x_{(n)})_{i_n,\overline{i_1\dots i_{n-1}i_{n+1}\dots i_N}}=x_{i_1,\dots,i_N}$.
\end{definition}
\begin{definition}[Outer Product]
    The \textit{outer product} of two vectors $\mathbf{a}\in\mathbb{R}^{I}$ and $\mathbf{b}\in\mathbb{R}^{J}$ is given by $\mathbf{c}=\mathbf{a}\circ\mathbf{b}\in\mathbb{R}^{I\times J}$, with $c_{i,j}=a_ib_j$.
\end{definition}
\begin{definition}[Kronecker Product]
    The \textit{Kronecker product} of two matrices $\mathbf{A}\in\mathbb{R}^{I\times J}$ and $\mathbf{B}\in\mathbb{R}^{K\times L}$ is denoted by $\mathbf{C}=\mathbf{A}\otimes \mathbf{B}\in\mathbb{R}^{IK\times JL}$, with $c_{(i-1)K+k,(j-1)L+l}=a_{i,j}b_{k,l}$.
\end{definition}
\begin{definition}[Khatri-Rao Product]
    The \textit{Khatri-Rao product} of two matrices $\mathbf{A}=[\mathbf{a}_1,\dots,\mathbf{a}_R]\in\mathbb{R}^{I\times R}$ and $\mathbf{B}=[\mathbf{b}_1,\dots,\mathbf{b}_R]\in\mathbb{R}^{J\times R}$ is denoted by $\mathbf{C}=\mathbf{A}\odot \mathbf{B}\in\mathbb{R}^{IJ\times R}$, where the columns $\mathbf{c}_r=\mathbf{a}_r\otimes \mathbf{b}_r$, $1\leq r\leq R$.
\end{definition}
\begin{definition}[Hadamard Product]
    The \textit{Hadamard product} of two $N$\textsuperscript{th}-order tensors, $\mathcal{A}\in\mathbb{R}^{I_1\times \dots \times I_N}$ and $\mathcal{B}\in\mathbb{R}^{I_1\times \dots \times I_N}$ is denoted by $\mathcal{C}=\mathcal{A}\circledast\mathcal{B}\in\mathbb{R}^{I_1\times \dots \times I_N}$, with $c_{i_1,\dots,i_N}=a_{i_1,\dots,i_N}b_{i_1,\dots,i_N}$.
\end{definition}
\begin{definition}[Tensor Contraction]
    The \textit{contraction} of an $N$\textsuperscript{th}-order tensor, $\mathcal{A}\in\mathbb{R}^{I_1\times \dots \times I_N}$, and an $M$\textsuperscript{th}-order tensor $\mathcal{B}\in\mathbb{R}^{J_1\times \dots \times J_M}$, over the $n$\textsuperscript{th} and $m$\textsuperscript{th} modes respectively, where $I_n=J_m$, results in an ($N+M-2$)\textsuperscript{th}-order tensor with entries $c_{i_1,\dots,i_{n-1},i_{n+1},\dots,i_N,j_1,\dots,j_{m-1},j_{m+1},\dots,j_M}=\sum_{i_n=1}^{I_n}a_{i_1,\dots,i_{n-1},i_n,i_{n+1},\dots,i_N}b_{j_1,\dots,j_{m-1},i_n,j_{m+1},\dots,j_M}$.
\end{definition}
\begin{definition}[Inner Product of Tensors]
    The \textit{inner product} of two $N$\textsuperscript{th}-order tensors $\mathcal{A}\in\mathbb{R}^{I_1\times \dots \times I_N}$ and $\mathcal{B}\in\mathbb{R}^{I_1\times \dots \times I_N}$ is denoted by $c=\langle\mathcal{A},\mathcal{B}\rangle\in\mathbb{R}$ with $c=\sum_{i_1,\dots,i_N}a_{i_1,\dots,i_N}b_{i_1,\dots,i_N}$.
\end{definition}

A property (see Appendix \ref{Khatri} for proof) used in this paper is
\begin{align}
    &\left(\mathbf{A}^{(1)}\odot\dots\odot \mathbf{A}^{(N)}\right)^T\left(\mathbf{B}^{(1)}\odot\dots\odot \mathbf{B}^{(N)}\right) \nonumber \\
    &=\mathbf{A}^{(1)T}\mathbf{B}^{(1)}\circledast\dots \circledast \mathbf{A}^{(N)T}\mathbf{B}^{(N)} \nonumber \\
    &=\bigcircledast_{k=1}^N\mathbf{A}^{(k)T}\mathbf{B}^{(k)}, \label{KhatriProp}
\end{align} 
where $\mathbf{A}^{(k)}\in\mathbb{R}^{I_k\times J}$ and $\mathbf{B}^{(k)}\in\mathbb{R}^{I_k\times L}$.

\subsection{Canonical Polyadic and Tensor Train Decompositions}
\subsubsection*{Tensor Networks}
A Tensor Network (TN) provides an efficient representation of a tensor through a set of lower-order tensors which are contracted over certain modes. The relatively low order of the core tensors within TNs and their sparse interconnections allow for the mitigation of the Curse of Dimensionality as the number of parameters within the TN representation tends to scale linearly with the tensor order (rather than exponentially as in the raw tensor format).

\subsubsection*{Canonical Polyadic Decomposition}
Canonical Polyadic Decomposition \cite{carroll1970analysis,harshman1970foundations} expresses a tensor $\mathcal{X}\in\mathbb{R}^{I_1\times\dots\times I_N}$ as a sum of outer products of vectors $\mathbf{a}^{(1)}_{r}, \mathbf{a}^{(2)}_{r}, \dots, \mathbf{a}^{(N)}_{r}$ (i.e., rank-1 terms):
\begin{align}
    \mathcal{X}&=\sum_{r=1}^{R}\mathbf{a}^{(1)}_{r}\circ\dots\circ \mathbf{a}^{(N)}_{r} 
    =\sum_{r=1}^{R}\bigcirc_{k=1}^N\mathbf{a}^{(k)}_{r},
\end{align}
where $R$ is the CP rank of the tensor (equal to the standard matrix rank when $N=2$). We can arrange these vectors into $N$ factor matrices $\mathbf{A}^{(n)}=[\mathbf{a}^{(n)}_1,\dots,\mathbf{a}^{(n)}_R]\in\mathbb{R}^{I_n\times R}$, ($1\leq n \leq N$), which allows the CP decomposition to be expressed as a contraction of the identity tensor, $\mathcal{I}\in\mathbb{R}^{I_1\times\dots\times I_N}$ (all ones on the super-diagonal) with the so-formed factor matrices. This view allows the CPD to be formulated as a TN, as is shown in the right panel in Fig. \ref{fig:cp_and_tt}.

Notice that the entries of the CPD can be expressed as
\begin{align}
    x_{i_1,\dots,i_N}&=\sum_{r=1}^R a^{(1)}_{i_1,r}a^{(2)}_{i_2,r}\cdots a^{(N)}_{i_N,r} \nonumber \\
                     &=\left(\bigcircledast_{k=1}^N\hat{\mathbf{a}}^{(k)}_{i_k}\right)\mathbf{1},
\end{align}
where $\hat{\mathbf{a}}^{(n)}_{j}$ denotes the $j$\textsuperscript{th} row of the $n$\textsuperscript{th} factor matrix, and $\mathbf{1}=[1,\dots,1]^T\in\mathbb{R}^{R}$.

Upon the matricization of the tensor $\mathcal{X}$, we obtain
\begin{IEEEeqnarray}{rCll}
    \mathbf{X}_{(n)}&=&\mathbf{A}^{(n)}\Big(\mathbf{A}^{(N)}\odot\dots \odot \mathbf{A}^{(n+1)}\odot &\mathbf{A}^{(n-1)} \nonumber \\
    &&& \odot \dots \odot \mathbf{A}^{(1)}\Big)^T \nonumber \\
     &=&\mathbf{A}^{(n)}\left(\bigodot_{\substack{k=N \\ k\neq n}}^1 \mathbf{A}^{(k)}\right)^T.&
\end{IEEEeqnarray}

Furthermore, the vectorization of $\mathcal{X}$ can be expressed as
\begin{align}
    \text{{\normalfont vec}}(\mathcal{X})=\left(\bigodot_{k=N}^1\mathbf{A}^{(k)}\right)\mathbf{1}.
\end{align}

\subsubsection*{Tensor Train Decomposition}
\begin{figure}
    \begin{subfigure}[b]{0.5\linewidth}
       \centering
       \includegraphics[scale=0.8]{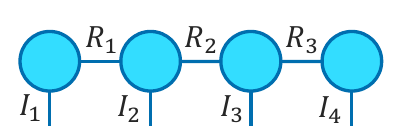}
    \caption{}
    \end{subfigure}\hfill
    \begin{subfigure}[b]{0.5\linewidth}
       \centering
       \includegraphics[scale=0.8]{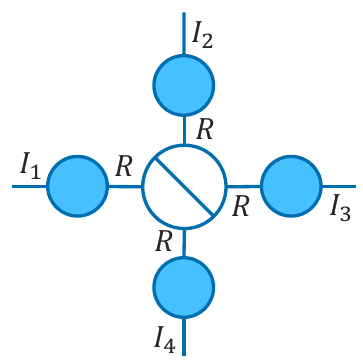}
    \caption{}
    \end{subfigure}

\caption{Tensor Network notation of an order-4 tensor. (a) Tensor Train and (b) Canonical Polyadic decomposition. A circle with a diagonal line denotes a super-diagonal tensor, in this case the identity tensor, with all unities on the diagonal.} \label{fig:cp_and_tt}
\end{figure}

A tensor $\mathcal{X}\in\mathbb{R}^{I_1\times\dots\times I_N}$ can be represented in the Tensor Train (TT) format as
\begin{align}
    \mathcal{X}=\mathcal{G}^{(1)}\times ^1\dots\times ^1 \mathcal{G}^{(N)},
\end{align}
where $\mathcal{G}^{(n)}\in\mathbb{R}^{R_{n-1}\times I_n \times R_{n}}$, $R_0=R_N=1$, are the TT cores, and $\mathcal{G}^{(n)}\times ^1 \mathcal{G}^{(n+1)}$ denotes contraction over the last mode of $\mathcal{G}^{(n)}$ and first mode of $\mathcal{G}^{(n+1)}$. The TT rank is given by the dimensions of the contracted modes, i.e., $\text{TT rank}=\{R_1,\dots,R_{N-1}\}$. The TT format is depicted in the left panel of Fig. \ref{fig:cp_and_tt}. As noted in \cite{zhao2016tensor}, an important drawback of TT is its sensitivity to the permutation of tensor dimensions.

For a more comprehensive review of other TDs and TNs, such as Tucker, Tensor Ring, and Hierarchical Tucker, we refer the reader to \cite{tensornetworksdimens,7038247,KoBa09}.

\section{CP-Based Predictor} \label{sec:cp_based_mod}
\subsection{Tensor Network Framework for Supervised Learning} \label{TN_framework}

\begin{figure}
    \begin{subfigure}[b]{0.5\linewidth}
       \centering
       \includegraphics[scale=0.8]{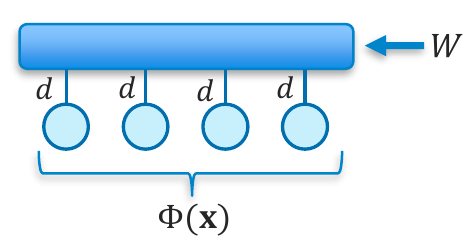}
    \caption{}
    \end{subfigure}\hfill
    \begin{subfigure}[b]{0.5\linewidth}
       \centering
       \includegraphics[scale=0.8]{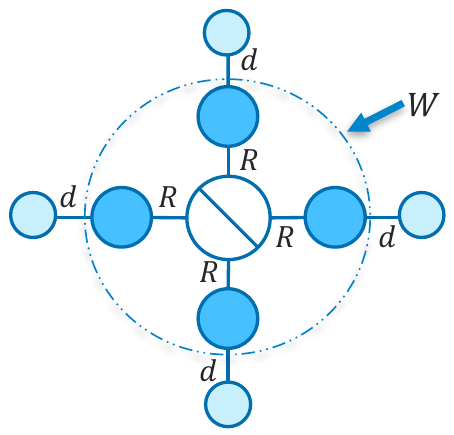}
    \caption{}
    \end{subfigure}          
\caption{Tensor-valued model prediction based on (\ref{inner_prod_pred}). (a) Without a tensor decomposition, (b) with CP decomposition of the weight tensor $\mathcal{W}$.} \label{fig:model_prediction}
\end{figure}

The framework for our work is based on \cite{NIPS2016_6211, novikov2018exponential}. Consider a supervised learning task where each sample is represented through a set of $N$ features. A \textit{local feature mapping} $\phi\colon \mathbb{R}\to\mathbb{R}^d$ is next applied to every feature $x_n$, where $d$ is referred to as the \textit{local dimension} of the mapping. The choice of the feature map is flexible and application-dependent. The outer product between the mapped feature vectors then yields
\begin{align}
    \Phi(\mathbf{x})=\bigcirc_{k=1}^N\phi(x_k)\in\mathbb{R}^{d^N}, \label{eq:pred}
\end{align}
where $\mathbb{R}^{d^N}$ denotes $\mathbb{R}^{\overbrace{d\times\dots\times d}^{N \text{times}}}$. For one output (e.g., single-target regression or binary classification), the prediction produced by the model in (\ref{eq:pred}) is given by
\begin{align}
    f(\mathbf{x})=\langle\Phi(\mathbf{x}),\mathcal{W}\rangle, \label{inner_prod_pred}
\end{align}
where $\mathcal{W}\in\mathbb{R}^{d^N}$ is the \textit{weight tensor}, which comprises all model coefficients. Note that in the case of multi-target regression or multi-class classification with $L$ labels, the model is composed of $L$ different weight tensors $\mathcal{W}^{l}$ ($1\leq l\leq L$). For simplicity and without loss of generality, we here consider only the single-target case, but the analysis can be easily generalized. For a graphical representation of the prediction in (\ref{inner_prod_pred}) in Tensor Network notation, see Figure \ref{fig:model_prediction}.
\begin{example} \label{ex:allcombs}
Consider the map $\phi$ such that $\phi(x_n)=\begin{bmatrix}
1 , x_n\end{bmatrix}^T$. Then, for the number of features $N=3$, we have $\Phi(\mathbf{x})\in \mathbb{R} ^{2\times 2\times 2} $ and
\begin{IEEEeqnarray}{rCl} 
     f(\mathbf{x})=w_{1,1,1}&&+w_{2,1,1}x_1+w_{1,2,1}x_2+w_{1,1,2}x_3+w_{2,2,1}x_1x_2 \nonumber \\
     &&+w_{2,1,2}x_1x_3+w_{1,2,2}x_2x_3+w_{2,2,2}x_1x_2x_3. \label{ex:eqs}
\end{IEEEeqnarray} 
Thus, the model captures all combinations of distinct features, $x_1, x_2, x_3$. 
\end{example}

Observe from (\ref{ex:eqs}) that the size of the weight tensor, $\mathcal{W}$, scales exponentially with the number of features and is therefore computationally prohibitive to learn. To this end, $\mathcal{W}$ can be represented as a Tensor Network \cite{NIPS2016_6211}, where the number of parameters scales linearly with the number of features. 

\subsection{Representing the Weight Tensor With CPD} \label{CP_frame}
In this work, we represent the weight tensor, $\mathcal{W}$, in the CP format, which, unlike the TT and other Tensor Network formats, is insensitive to the ordering of the features. Although this also holds true for the Tucker decomposition \cite{7038247}, using the Tucker format comes at a cost of exponential scaling with the number of features. Also, the Tucker decomposition involves a core tensor, the modes of which typically have dimensions smaller than those of the original tensor, thus rendering the Tucker decomposition unsuitable for weight tensors with dimensions 2 across all modes, such as in Example \ref{ex:allcombs}.

It is crucial to note that by virtue of the rank-1 structure of the implicit tensor of feature interactions coupled with the assumed CP format of the weight tensor, in this work \textit{we do not deal with higher-order tensors, no decomposition is actually performed, and thus no alternating least-squares algorithm is required}. Instead, all computations can be efficiently performed with relatively low-dimensional vectors and matrices, and we can use stochastic gradient descent to fit the model (see Section \ref{Algorithm}). Furthermore, since a tensor decomposition is not our end-goal, determining the \textit{optimal} rank (which is known to be difficult for CPD) of the implicit weight tensor is not an issue here. We therefore treat the rank of the weight tensor simply as a 1-D hyperparameter and, as shown in our experiments, performance is not sensitive to the rank for all but very small values (see Fig. \ref{fig:ranks}). 


Another well-known issue associated with the CP decomposition is that of degeneracy \cite{mitchell1994slowly}. Namely, in some scenarios it is possible for the norms of the rank-1 terms of the CP decomposition to become arbitrarily large while still reducing the approximation error by cancelling each other. Although, as discussed, we never actually decompose a tensor, it is still possible for the parameters of our implicit tensor to become large during training. This can be counteracted  by simply using L2 regularization, though our model obtained good predictive performance even without it. 

\section{Efficient Algorithms for Prediction} \label{Algorithm}

\subsection{Prediction Algorithm}
Given the factor matrices, $\mathbf{A}^{(n)}$, and the map $\phi$, there are various ways of obtaining model predictions, some of which are dramatically more computationally efficient than others. For example, a direct implementation of (\ref{inner_prod_pred}) would yield a procedure that scales exponentially with the model order $N$.

To arrive at an efficient procedure for computing the model prediction\footnote{We refer the reader to the graphical tensor notation in this article to gain further intuition into the underlying algorithms.}, we start by expressing it as 
\begin{align}
    f(\mathbf{x})&=\langle \Phi(\mathbf{x}),\mathcal{W}\rangle \nonumber \\
                 &=\langle \text{vec}\left(\Phi(\mathbf{x})\right),\text{vec}\left(\mathcal{W}\right)\rangle \nonumber \\
                 &=\text{vec}\left(\Phi(\mathbf{x})\right)^T \text{vec}\left(\mathcal{W}\right) \nonumber \\
                 &=\left(\bigodot_{k=N}^1\phi(x_k)\right)^T\left(\bigodot_{k=N}^1\mathbf{A}^{(k)}\right)\mathbf{1} \nonumber \\
                 &=\left(\bigcircledast_{k=1}^{N}\phi^T(x_k)\mathbf{A}^{(k)}\right)\mathbf{1}. \label{prediction}
\end{align}

\begin{figure}
\centering
     \includegraphics[scale=0.7]{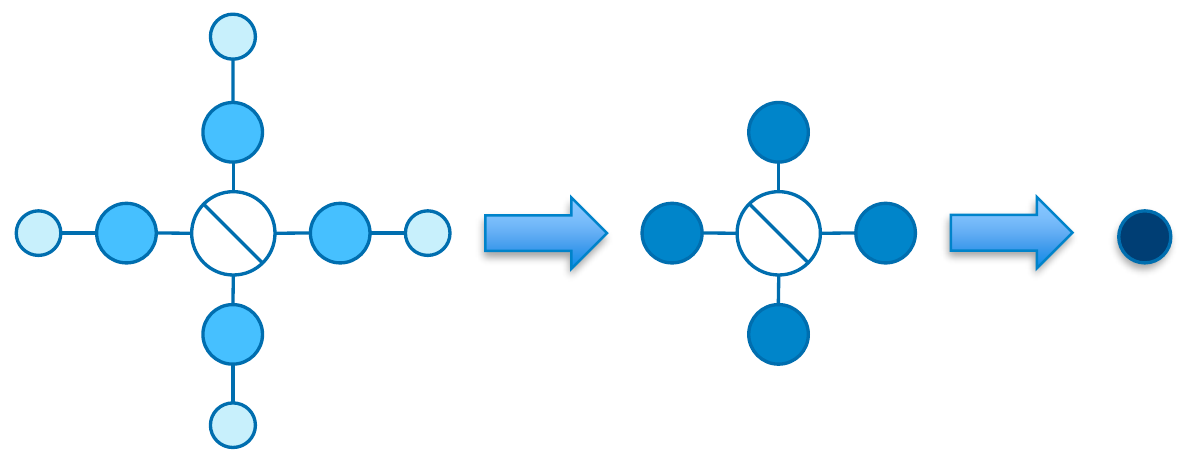}
     \caption{Model prediction (\ref{prediction}) in graphical TN notation, which amounts to contracting the TN in Fig. \ref{fig:model_prediction}. The first arrow denotes matrix-vector multiplications for all factor matrices, and the second arrow the sum of Hadamard products of all resulting vectors.}
     \label{fig:cpd_predict}
\end{figure}

The corresponding computation procedure is given in Algorithm \ref{algo:pred}, the complexity\footnote{Throughout the paper we give the asymptotic complexity of the algorithms, assuming that each operation is executed in sequence. The methods are, however, highly amenable to parallelization, and so they can be implemented in e.g., \texttt{TensorFlow} very efficiently.} of which is $\mathcal{O}(NRd)$. Figure \ref{fig:cpd_predict} shows the graphical visualization of the proposed algorithm. Notice that obtaining the model predictions amounts to contracting the TN in Fig. \ref{fig:model_prediction}.

\begin{algorithm}
\SetAlgoLined
\textbf{Input}: Data point $\mathbf{x}\in\mathbb{R}^N$ and factor matrices $\mathbf{A}^{(n)}\in\mathbb{R}^{d\times R}$, $1\leq n \leq N$ \\
\textbf{Output}: Prediction $\hat{y}\in\mathbb{R}$ \\
\Begin{
 \tcp{Construct $\phi(x_n)\in\mathbb{R}^{d\times 1}$ for $1\leq n\leq N$}
 $\mathbf{p}=\mathbf{1}^T\in\mathbb{R}^{1\times R}$ \tcp{Initialize (row) vector of ones}
 \For{$n=1,\dots,N$}{
  $\mathbf{p}\leftarrow \mathbf{p}\circledast \phi^T(x_n)\mathbf{A}^{(n)}$
 }
 $\hat{y}=\text{sum}(\mathbf{p})$ \tcp{Sum entries of $\mathbf{p}$}
 }
 \caption{Model Prediction}
 \label{algo:pred}
\end{algorithm}

\begin{figure} 
\centering
     \includegraphics[scale=0.7]{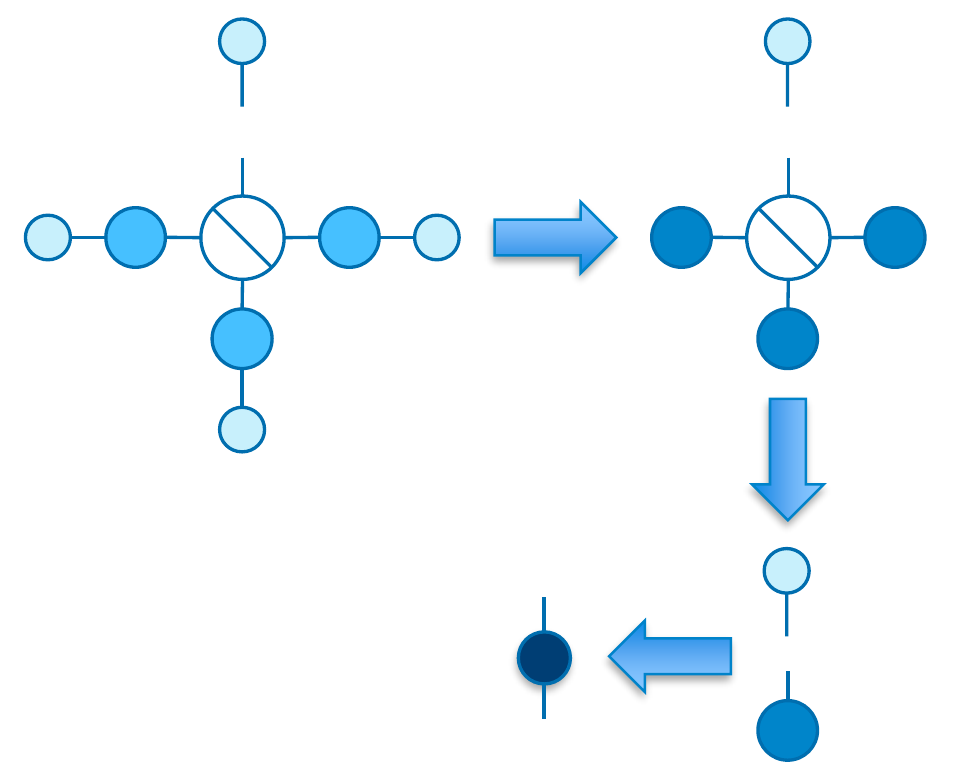}
     \caption{Partial derivative of prediction in (\ref{prediction}) w.r.t. a given factor matrix. The first step corresponds to matrix-vector multiplications for all but one factor matrices, the second step to the Hadamard product of the resulting vectors, and the third to the outer product of the remaining vectors.}
     \label{fig:learning_algo}
\end{figure}

\subsection{Learning Algorithm}
In order to learn the model parameters (i.e., factor matrices) using an approach based on first-order derivatives (e.g., stochastic gradient descent, ADAM \cite{adam}, etc.) we need to obtain the partial derivative of the prediction (model output) with respect to each factor matrix (the desired gradient can subsequently be computed using the chain rule). Alternatively, one can employ automatic differentiation, for example as in Keras \cite{chollet2015keras}, and specify only the forward pass (i.e., Algorithm \ref{algo:pred}).

We derive this partial derivative analytically using the mode-$n$ matricization of tensors, as follows:
\begin{align}
    f(\mathbf{x})&=\langle \Phi(\mathbf{x}),\mathcal{W}\rangle \nonumber \\
                 &=\langle \mathbf{\Phi(\mathbf{x})}_{(n)},\mathbf{W}_{(n)}\rangle \nonumber \\
                 &=\Tr\left(\mathbf{\Phi}^T(\mathbf{x})_{(n)}\mathbf{W}_{(n)}\right) \nonumber \\
                 &=\Tr\left(\left(\bigodot_{\substack{k=N \\ k\neq n}}^1 \phi(x_k)\right)\phi^T(x_n)\mathbf{A}^{(n)}\left(\bigodot_{\substack{k=N \\ k\neq n}}^1 \mathbf{A}^{(k)}\right)^T\right) \nonumber \\
                 &=\Tr\left(\mathbf{A}^{(n)}\left(\bigodot_{\substack{k=N \\ k\neq n}}^1 \mathbf{A}^{(k)}\right)^T\left(\bigodot_{\substack{k=N \\ k\neq n}}^1 \phi(x_k)\right)\phi^T(x_n)\right) \nonumber \\
                 &=\Tr\left(\mathbf{A}^{(n)}\left(\bigcircledast_{\substack{k=1 \\ k\neq n}}^N\mathbf{A}^{(k)T} \phi(x_k)\right)\phi^T(x_n)\right),
\end{align}
where $\Tr(\cdot)$ denotes the trace operator of which its cyclic property is used here.

The partial derivative of the prediction w.r.t. the $n$\textsuperscript{th} factor matrix $ \mathbf{A}^{(n)}$ is then given by
\begin{align} 
    \frac{\partial f(\mathbf{x})}{\partial \mathbf{A}^{(n)}}=\phi(x_n)\left(\bigcircledast_{\substack{k=1 \\ k\neq n}}^N\phi^T(x_k)\mathbf{A}^{(k)}\right), \label{eq:part_deriv}
\end{align}
where we have used the well-known matrix calculus rule
\begin{align}
     \frac{\partial}{\partial \mathbf{X}}\Tr\left(\mathbf{X}\mathbf{W}\right)=\mathbf{W}^T.
\end{align}

The corresponding procedure is depicted in graphical notation in Fig. \ref{fig:learning_algo}. Notice that a naive implementation of (\ref{eq:part_deriv}) could lead to a computational complexity of $\mathcal{O}(NRd)$ for obtaining the partial derivative w.r.t. each factor matrix and thus, a total of $\mathcal{O}(N^2Rd)$ w.r.t. to all factors. The quadratic scaling with $N$ is due to the repetition of the same matrix-vector products as we take the derivative w.r.t. different factors, which can be avoided by storing these products. This leads to the proposed Algorithm \ref{algo:learning}, which has a complexity of $\mathcal{O}(NRd)$.

\begin{algorithm}
\SetAlgoLined
\textbf{Input}: Data point $\mathbf{x}\in\mathbb{R}^N$ and factor matrices $\mathbf{A}^{(n)}\in\mathbb{R}^{d\times R}$, $1\leq n \leq N$ \\
\textbf{Output}: Partial derivative $\frac{\partial f(\mathbf{x})}{\partial \mathbf{A}^{(n)}}\in\mathbb{R}^{d\times R}$ for $1\leq n \leq N$ \\
\Begin{
\tcp{Construct $\phi(x_n)\in\mathbb{R}^{d\times 1}$ for $1\leq n \leq N$}
$\mathbf{p}=\mathbf{1}^T\in\mathbb{R}^{1\times R}$ \tcp{Initialize (row) vector of ones}
 \For{$n=1,\dots,N$}{
 $\mathbf{m}_n=\phi^T(x_n)\mathbf{A}^{(n)}\in\mathbb{R}^{1\times R}$ \tcp{Store matrix-vector products}
 $\mathbf{p}\leftarrow\mathbf{p}\circledast \mathbf{m}_n$ \tcp{Store Hadamard products}
 }
 \For{$n=1,\dots,N$}{
  $\mathbf{d}=\mathbf{p}\oslash \mathbf{m}_n\in\mathbb{R}^{1\times R}$ \tcp{Divide element-wise}
  $\frac{\partial f(\mathbf{x})}{\partial \mathbf{A}^{(n)}}=\phi(x_n) \mathbf{d}$
 }
 }
 \caption{Partial Derivative of Prediction}
 \label{algo:learning}
\end{algorithm}

\section{Local Feature Mappings} \label{sec:feat_mappings}
The choice of feature map $\phi$ underpins our model, and here we propose feature maps for both dense and sparse data.
\subsection{Polynomial Mapping}
For generality, for dense data, we propose features maps of the form
\begin{align} \label{eq:poly_mapping}
    \phi_d(x_n)=[1,x_n,x_n^2,\dots,x_n^{(d-1)}]^T,
\end{align}
which is a higher dimensional generalization of the map used in previous works such as in \cite{novikov2018exponential}, i.e., $\phi_{\text{EM}}(x_n)=[1,x_n]^T$. The subscript in (\ref{eq:poly_mapping}) is used to designate the dependence on the local feature dimension $d$. From (\ref{eq:poly_mapping}), it is important to notice that it:
\begin{itemize}
    \item Comprises the linear model (see Section \ref{sec:linear_model_init});
    \item Allows for a straightforward modification of the local dimension $d$, which gives us additional degrees of freedom (aside from the rank $R$) to vary the expressiveness of the model.
\end{itemize}
\subsection{Normalized Polynomial Mapping}
For a high enough $d$, the feature map in (\ref{eq:poly_mapping}) exhibits instability in the training process and very high overfitting, especially in the presence of outliers. As a remedy to this issue, we propose to normalize the resulting vector to unit length; this enables us to use a very high $d$, even $d>100$, without numerical issues. As shown in Section \ref{sec:experiments}, such normalization leads to a significantly better performance in the experiments conducted. 

The proposed normalized map can be expressed as
\begin{align} \label{eq:unit_norm_vec}
    \hat{\phi}_d(x_n)=\frac{1}{\sqrt{\sum_{k=0}^{d-1}x_n^{2k}}}[1,x_n,x_n^2,\dots,x_n^{(d-1)}]^T
\end{align}
Note that, due to the dependence of the denominator on the feature values, a bias term is no longer present. Should we desire to involve bias, this may be achieved by considering a bias as an independent parameter; we achieved best performance in our experiments by simply using (\ref{eq:unit_norm_vec}).

As is customary in data analytics, the data are pre-processed by standardizing the features; otherwise, an increase in $d$ would quickly result in numerical overflow. Using a standard deviation of 1 and mean of 0 places almost all feature values in the range between -3 and 3, and so, in our experiments, even $d=75$ did not cause overflow for a 32-bit single-precision floating-point representation (one can use double-point precision for extremely high $d$). Furthermore, the entries of the normalized vector in (\ref{eq:unit_norm_vec}) either decrease exponentially with the index if the standardized feature values are between -1 and 1, or increase exponentially otherwise. Hence, after normalization, only a few of the vector entries have a non-negligible impact on the prediction.

To gain further intuition into how the normalized map facilitates smooth learning curves, consider an upper bound on the Frobenius norm of the partial derivative of the prediction w.r.t. each factor matrix when $\norm{\phi(x_n)}=1$:
\begin{align}
    \norm{\frac{\partial f(\mathbf{x})}{\partial \mathbf{A}^{(n)}}}&\leq \norm{\bigcircledast_{\substack{k=1 \\ k\neq n}}^N\phi^T(x_k)\mathbf{A}^{(k)}}\leq \prod_{\substack{k=1 \\ k\neq n}}^N\norm{\phi^T(x_k)\mathbf{A}^{(k)}} \nonumber \\
    &\leq \prod_{\substack{k=1 \\ k\neq n}}^N\norm{\mathbf{A}^{(k)}}.
\end{align}
Therefore, the norm of this derivative is upper bounded by a value that does not depend on $\phi(x_n)$. Since the absolute value of the prediction $f(\mathbf{x})$ is bounded in a similar way, each step of gradient descent (when using e.g., mean squared error as the loss function) is also bounded. Constraining the norm of $\phi(x_n)$ prevents very large steps that could otherwise occur with high $d$ and feature values greater than 1.  

\subsection{Mapping for Categorical (Sparse) Data} \label{subsec:map_cat}

An advantage of the proposed framework for supervised learning is its ability to handle categorical features after they have been one-hot encoded. Although it is possible to concatenate all binary one-hot features into a large feature vector and allocate a factor matrix to each feature (thereby enabling the use of (\ref{eq:poly_mapping}) with $d=2$), this would unnecessarily model interactions between one-hot features belonging to the same categorical variable. To this end, as proposed in \cite{novikov2018exponential}, we employ the feature map for categorical data given by
\begin{align}
\phi(x_n)=\left[1,\mathbf{v}_n^T\right]^T,    
\end{align}
where $\mathbf{v}_n\in\mathbb{R}^{K_n}$ is the one-hot-encoded representation of the $n$\textsuperscript{th} categorical feature, and $K_n$ the number of values that the feature can assume. 

\section{Order Regularization for CP-Based Predictor} \label{Regularization}
Although the CP format itself provides strong regularization for small rank\footnote{The main assumption is that the true matrices we aim to approximate lie near some low-dimensional subspaces.}, there is room to improve the performance by guiding the training algorithm to hypotheses that are more likely to generalize well for unseen data. However, standard regularizers, such as those based on L1 and L2 norms, cannot differentiate between the coefficients for different feature interactions in (\ref{eq:mod_pred}). This is likely to prove suboptimal when using the polynomial or categorical mappings, as often the application dictates to constrain the coefficients of the higher-order terms relatively more. \textit{Order regularization} \cite{novikov2018exponential} addresses this issue by penalizing large coefficients for higher-order terms more than those for lower-order ones. We now proceed to derive an efficient order regularization for the CP-based predictor.

The penalty for order regularization is given by $\langle\mathcal{B}\circledast\mathcal{W},\mathcal{B}\circledast\mathcal{W}\rangle$, where $\mathcal{B}=\bigcirc_{k=1}^N \mathbf{b}$ for a user-defined vector $\mathbf{b}$. In the case of polynomial functions, one choice is $\mathbf{b}=[1,\beta,\beta^2,\dots,\beta^{d-1}]^T$ with $\beta>1$. In other words, the coefficients of the higher-order terms are multiplied by a higher power of $\beta$ in the computation of the penalty function, and the factor matrices are adjusted in such a way so as to shrink these coefficients relatively more. In the case of categorical feature mapping, $\mathbf{b}=[1,\beta,\beta,\dots,\beta]^T$ is more suitable, since all binary, one-hot features should be treated equally.
\subsection{Computation of the Order Regularization Penalty}

\begin{figure} 
\centering
     \includegraphics[scale=0.7]{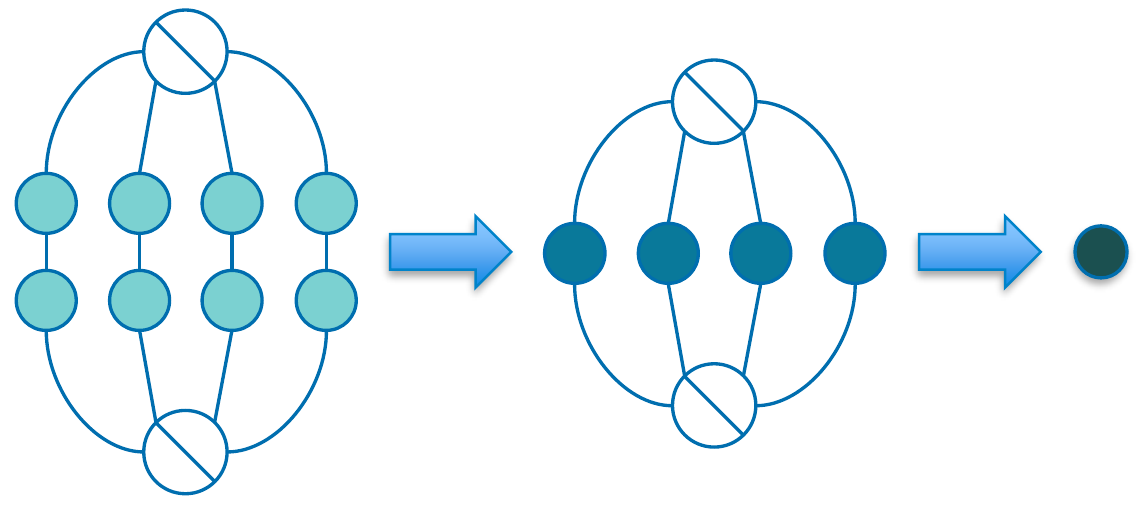}
     \caption{Order regularization penalty in graphical TN notation for $N=4$. The first arrow denotes matrix-matrix multiplications for all factor matrices and second arrow designates the sum of Hadamard products of all resulting matrices. Note that the matrices in the left-most TN result from the Hadamard products $\mathbf{A}^{(n)}\circledast\mathbf{B}$.}
     \label{fig:order_reg_pen_diag}
\end{figure}

A closer inspection of the term $\mathcal{B}\circledast\mathcal{W}$ gives

\begin{align}
    \mathcal{B}\circledast\mathcal{W}&=\left(\bigcirc_{k=1}^N\mathbf{b}\right)\circledast \left(\sum_{r=1}^{R}\bigcirc_{k=1}^N\mathbf{a}^{(k)}_{r}\right) \nonumber \\
    &=\sum_{r=1}^{R}\bigcirc_{k=1}^N\left(\mathbf{a}^{(k)}_{r}\circledast\mathbf{b}\right).
\end{align}
Now, let $\mathbf{B}=\left[\mathbf{b},\dots,\mathbf{b}\right]\in\mathbb{R}^{d\times R}$ and $\mathbf{Y}^{(n)}=\mathbf{A}^{(n)}\circledast\mathbf{B}$. The penalty $\langle\mathcal{B}\circledast\mathcal{W},\mathcal{B}\circledast\mathcal{W}\rangle$ can now be re-written as
\begin{align}
    P(\mathcal{B},\mathcal{W})&=\text{vec}\left(\mathcal{B}\circledast\mathcal{W}\right)^T\text{vec}\left(\mathcal{B}\circledast\mathcal{W}\right) \nonumber \\
        &=\mathbf{1}^T\left(\bigodot_{k=N}^1\mathbf{Y}^{(k)}\right)^T\left(\bigodot_{k=N}^1\mathbf{Y}^{(k)}\right)\mathbf{1} \nonumber \\
        &=\mathbf{1}^T\left(\bigcircledast_{k=1}^{N}\mathbf{Y}^{(k)T}\mathbf{Y}^{(k)}\right)\mathbf{1}. \label{eq:order_reg}
\end{align}
The procedure for computing the regularization penalty is given in Algorithm \ref{algo:order_reg} and depicted graphically in Fig. \ref{fig:order_reg_pen_diag}. The algorithm has a computational cost of $\mathcal{O}(NR^2d)$, so that the training procedure now scales quadratically in $R$ with order regularization (rather than linearly as before).

\begin{algorithm} 
\SetAlgoLined
\textbf{Input}: Factor matrices $\mathbf{A}^{(n)}\in\mathbb{R}^{d\times R}$, $1\leq n\leq N$, order regularization parameter $\mathbf{b}\in\mathbb{R}^{d\times 1}$, and regularization constant $\alpha\in\mathbb{R}$ \\
\textbf{Output}: Penalty $P\in\mathbb{R}$ \\
\Begin{
 $\mathbf{B}=[\mathbf{b},\dots,\mathbf{b}]^T\in\mathbb{R}^{d\times R}$ \\
 $\mathbf{P}=[\mathbf{1},\dots,\mathbf{1}]^T\in\mathbb{R}^{R\times R}$ \tcp{Initialize all-ones matrix}
 \For{$n=1,\dots,N$}{
  $\mathbf{P}\leftarrow \mathbf{P}\circledast\left( \left(\mathbf{A}^{(n)}\circledast\mathbf{B}\right)^T\left(\mathbf{A}^{(n)}\circledast\mathbf{B}\right)\right)$
 }
 $P=\alpha*\text{sum}(\mathbf{P})$ \tcp{Sum entries of $\mathbf{P}$}
 }
 \caption{Order Regularization Penalty}
 \label{algo:order_reg}
\end{algorithm}

\subsection{Partial Derivative of the Order Regularization Penalty}
The partial derivative of the order regularization penalty w.r.t. each factor matrix can be found by first expressing the penalty in the following way:
\begin{IEEEeqnarray}{rCl}
    P&=&\left\langle \left(\mathbf{B}_{(n)}\circledast\mathbf{W}_{(n)}\right),\left(\mathbf{B}_{(n)}\circledast\mathbf{W}_{(n)}\right) \right\rangle \nonumber \\
    &=&\Tr\left(\left(\mathbf{B}_{(n)}\circledast\mathbf{W}_{(n)}\right)^T\left(\mathbf{B}_{(n)}\circledast\mathbf{W}_{(n)}\right)\right) \nonumber \\
    &=&\Tr\left(\left(\bigodot_{\substack{k=N \\ k\neq n}}^1 \mathbf{Y}^{(k)}\right)\mathbf{Y}^{(n)T}\mathbf{Y}^{(n)}\left(\bigodot_{\substack{k=N \\ k\neq n}}^1 \mathbf{Y}^{(k)}\right)^T\right) \nonumber \\
    &=&\Tr\left(\mathbf{Y}^{(n)}\left(\bigodot_{\substack{k=N \\ k\neq n}}^1 \mathbf{Y}^{(k)} \right)^T\left(\bigodot_{\substack{k=N \\ k\neq n}}^1 \mathbf{Y}^{(k)}\right)\mathbf{Y}^{(n)T}\right) \nonumber \\
    &=&\Tr\left(\mathbf{Y}^{(n)}\left(\bigcircledast_{\substack{k=1 \\ k\neq n}}^N \mathbf{Y}^{(k)T}\mathbf{Y}^{(k)}\right)\mathbf{Y}^{(n)T}\right).
\end{IEEEeqnarray}
Then, using the identity (see Appendix \ref{Proof} for proof)
\begin{IEEEeqnarray}{rCl}
    \frac{\partial}{\partial \mathbf{X}}\Tr\Big((\mathbf{X}\circledast\mathbf{W})\mathbf{Z}&&(\mathbf{X}\circledast\mathbf{W})^T\Big)= \nonumber \\
    && \mathbf{W}\circledast\Big(\big(\mathbf{X}\circledast\mathbf{W}\big) \big(\mathbf{Z}+\mathbf{Z}^T\big)\Big) \qquad \label{eq:mat_calc_identity}
\end{IEEEeqnarray}
we arrive at (notice that in this case $\mathbf{Z}=\mathbf{Z}^T$)
\begin{align}
    \frac{\partial P}{\partial \mathbf{A}^{(n)}}=2\mathbf{B}\circledast\left(\mathbf{Y}^{(n)}\left(\bigcircledast_{\substack{k=1 \\ k\neq n}}^N\mathbf{Y}^{(k)T}\mathbf{Y}^{(k)}\right)\right).
\end{align}
As in Algorithm \ref{algo:learning}, by storing the Hadamard products, the procedure for computing the partial derivative of the penalty w.r.t. all factor matrices scales linearly with $N$ rather than quadratically, leading to a time complexity of $\mathcal{O}(NR^2d)$ (see Algorithm \ref{algo:order_reg_deriv}); the corresponding memory complexity is $\mathcal{O}(NRd+R^2)$. To further speed up the algorithm, it is possible to store the matrix products $\left(\mathbf{A}^{(n)}\circledast\mathbf{B}\right)^T\left(\mathbf{A}^{(n)}\circledast\mathbf{B}\right)$ in the first \texttt{for} loop to avoid their computation in the second one. This, however, would have no effect on the asymptotic computational complexity and would increase the memory complexity to $\mathcal{O}(NRd+NR^2)$.
\begin{algorithm}
\SetAlgoLined
\textbf{Input}: Factor matrices $\mathbf{A}^{(n)}\in\mathbb{R}^{d\times R}$, $1\leq n \leq N$, order regularization parameter $\mathbf{b}\in\mathbb{R}^{d\times 1}$, and regularization coefficient $\alpha\in\mathbb{R}$ \\
\textbf{Output}: Partial derivative $\frac{\partial P}{\partial \mathbf{A}^{(n)}}\in\mathbb{R}^{d\times R}$ for $1\leq n \leq N$ \\
\Begin{
$\mathbf{B}=[\mathbf{b},\dots,\mathbf{b}]^T\in\mathbb{R}^{d\times R}$ \\
 $\mathbf{P}=[\mathbf{1},\dots,\mathbf{1}]^T\in\mathbb{R}^{R\times R}$ \tcp{Initialize all-ones matrix}
 \For{$n=1,\dots,N$}{
 $\mathbf{P}\leftarrow \mathbf{P}\circledast\left( \left(\mathbf{A}^{(n)}\circledast\mathbf{B}\right)^T\left(\mathbf{A}^{(n)}\circledast\mathbf{B}\right)\right)$ \tcp{Store Hadamard products}
 }
 \For{$n=1,\dots,N$}{
  $\mathbf{D}=\mathbf{P}\oslash \left( \left(\mathbf{A}^{(n)}\circledast\mathbf{B}\right)^T\left(\mathbf{A}^{(n)}\circledast\mathbf{B}\right)\right)\in\mathbb{R}^{R\times R}$ \tcp{Divide element-wise}
  $\frac{\partial P}{\partial \mathbf{A}^{(n)}}=2\alpha\mathbf{B}\circledast\left(\left(\mathbf{A}^{(n)}\circledast\mathbf{B}\right)\mathbf{D}\right)$
 }
 }
 \caption{Partial Derivative of Order Regularization Penalty}
 \label{algo:order_reg_deriv}
\end{algorithm}

\section{Initialization of Factor Matrices} \label{sec:linear_model_init}
One way to initialize the factor matrices is to use independent zero-mean Gaussian noise, with a tunable standard deviation. However, if the local feature map is of the form $\phi(x_n)=\left[1,\psi^{(1)}(x_n),\dots,\psi^{(d-1)}(x_n)\right]^T$, where $\psi^{(j)}\colon \mathbb{R}\to\mathbb{R}$, it is possible to initialize the factors matrices by employing the linear model solution (linear or logistic regression depending on the task), trained on the set of features $\{\psi^{(j)}(x_n)\mid 1\leq j\leq d-1, 1\leq n\leq N\}$. Note that the linear model can also be trained on a subset of this set, as is performed in Example \ref{ex:lin_init}, but for clarity we assume for now the full set of transformed features.

Let $b$ denote the bias term of the linear model trained on the aforementioned set; in addition, let $w_{n,j}$ denote its weight corresponding to the $n$\textsuperscript{th} feature and the function $\psi^{(j)}$. Then, the CP-based predictor produces the same predictions as the linear model if the entries of the factor matrices are given by
\begin{align}
    a_{1,r}^{(n)}&=
    \begin{cases}
    \frac{b}{N}, & \text{{\normalfont for }} r=n, \\
    1,           & \text{{\normalfont for }} 1\leq r\leq N, r\neq n \\
    0,            & \text{{\normalfont otherwise}}.
    \end{cases}, \nonumber \\
    a_{j,r}^{(n)}&=
    \begin{cases}
    w_{r,j-1}, & \text{{\normalfont for }} r=n, j=2,\dots,d, \\
    0,           & \text{{\normalfont otherwise}}.
    \end{cases} \label{eq:lin_init_rows}
\end{align}
With such an initialization, the error at the first epoch tends to be lower than with random initialization and sometimes also converges to a lower value. This is especially common when the number of features is high (e.g., larger than 20). 

To prove that the initialization indeed yields the same prediction as the linear model, notice from (\ref{eq:mod_pred}) that the bias term for the CP-based predictor is obtained through the sum of the Hadamard products of the first rows of the factor matrices, to give 
\begin{align}
    \left(\bigcircledast_{k=1}^N\hat{\mathbf{a}}^{(k)}_{1}\right)\mathbf{1}=
    \begin{bmatrix}\smash[b]{\underbrace{\begin{matrix}\frac{b}{N},\dots,\frac{b}{N}\end{matrix}}_{N \text{ times}}},0,\dots,0
    \end{bmatrix}\mathbf{1}=b.
\end{align}
Moreover, the coefficient for $\psi_{j}(x_n)$ is obtained as
\begin{IEEEeqnarray}{rCl}
    \left(\bigcircledast_{\substack{k=1 \\ k\neq n}}^N\hat{\mathbf{a}}_1^{(k)}\circledast\right.&&\left. \hat{\mathbf{a}}_{j+1}^{(n)}\sizecorr{\bigcircledast_{\substack{k=1 \\ k\neq n}}^N}\right)\mathbf{1}=\left(\begin{bmatrix}\smash[b]{\underbrace{\begin{matrix}\frac{b}{N},\dots,\frac{b}{N}\end{matrix}}_{n-1 \text{ times}}},1,\smash[b]{\underbrace{\begin{matrix}\frac{b}{N},\dots,\frac{b}{N}\end{matrix}}_{N-n \text{ times}}},0,\dots,0
    \end{bmatrix}\right. \nonumber \\ \nonumber\\
    &&\left.\circledast \begin{bmatrix}\smash[b]{\underbrace{\begin{matrix}0,\dots,0\end{matrix}}_{n-1 \text{ times}}},w_{n,j},0,\dots,0
    \end{bmatrix}\right)\mathbf{1}=w_{n,j}. \\ \nonumber
\end{IEEEeqnarray}
while the coefficient for the second-order interaction $\psi_q(x_l)\psi_r(x_m)$ for $1\leq q,r\leq d-1$, $1\leq l,m \leq N$, and $l\neq m$ becomes
\begin{IEEEeqnarray}{rCl}
    &&\left(\bigcircledast_{\substack{k=1 \\ k\neq l \\ k\neq m}}^N\hat{\mathbf{a}}_1^{(k)}\circledast \hat{\mathbf{a}}_{q+1}^{(l)}\circledast\hat{\mathbf{a}}_{r+1}^{(m)}\right)\mathbf{1}=0
\end{IEEEeqnarray}
since $\hat{\mathbf{a}}_{q+1}^{(l)}\circledast\hat{\mathbf{a}}_{r+1}^{(m)}=\mathbf{0}$ for all $l$, $m$, $l\neq m$ and all $q$, $m$. Finally, in the same spirit, it is obvious that the coefficients of the higher-order interactions are also 0.

\begin{example} \label{ex:lin_init}
A consequence of (\ref{eq:lin_init_rows}) is that the CP-based predictor with the polynomial feature map, given in (\ref{eq:poly_mapping}), inherently contains the solution of the linear model trained on the original features. Namely, given the coefficient for the $n$\textsuperscript{th} feature $w_n$, the relevant initialization for the entries of the factor matrices is given by
\begin{align}
    a_{1,r}^{(n)}&=
    \begin{cases}
    \frac{b}{N}, & \text{{\normalfont for }} r=n, \\
    1,           & \text{{\normalfont for }} 1\leq r\leq N, r\neq n \\
    0,            & \text{{\normalfont otherwise}},
    \end{cases}, \nonumber \\
    a_{2,r}^{(n)}&=
    \begin{cases}
    w_n, & \text{{\normalfont for }} r=n, \\
    0,            & \text{{\normalfont otherwise}},
    \end{cases}
\end{align}
and all other rows equal $\mathbf{0}$, since they correspond to features raised to powers higher than 1.
\end{example}

\begin{remark}
When dealing with categorical features that are one-hot encoded, the initialization changes slightly due to the fact that the number of rows of the factor matrices is not constant (see Section \ref{sec:feat_mappings}). Given $N$ categorical features, each assuming $K_n$ ($1\leq n\leq N$) values, the rows of the factor matrices can be initialized as in (\ref{eq:lin_init_rows}), except that $d$ must be replaced with $K_n$. Note that in this case, $w_{n,j}$ denotes the weight of the binary feature corresponding to the $n$\textsuperscript{th} categorical feature and its $j$\textsuperscript{th} possible value.
\end{remark}

\section{Numerical Experiments} \label{sec:experiments}

To demonstrate the generality and flexibility of the proposed approach, comprehensive experiments were performed over three case studies of different nature:

\begin{itemize}
    \item A synthetic polynomial regression task, in order to  illustrate the effects of initialization from the linear model solution and order regularization, for lower than and higher than optimal local dimensions;
    \item MovieLens 100K recommender system dataset (classification on very sparse categorical data);
    \item The California Housing dataset (regression on dense data), to show the utility of using the introduced very high local dimension, coupled with normalized polynomial feature mappings.

\end{itemize}
The MovieLens and California housing datasets were chosen as illustrative benchmark examples of very common tasks in machine learning.

We evaluated the performance of our proposed CP-based predictor against the existing TT-based predictor using the code provided in Google's TensorNetwork library \cite{DBLP:journals/corr/abs-1906-06329} for the latter. To further demonstrate the superiority of the proposed CP-based predictor over other TN-based predictors for non-sequential data, we implemented the Tucker-, Tensor Ring-, and Hierarchical Tucker-based predictors. We also compare with other popular general predictors. We used the \texttt{scikit-learn} implementation for SVM and the \texttt{Keras} library \cite{chollet2015keras} to construct the neural networks, as well as \texttt{polylearn}\footnote{https://github.com/scikit-learn-contrib/polylearn} and \texttt{tffm}\footnote{https://github.com/geffy/tffm} for the Polynomial Networks and Higher-Order Factorization Machines implementations, respectively.

\subsection{Effects of Initialization and Order Regularization} \label{subsec:init_reg}
A synthetic dataset was used for this experiment, with the target function set to a 2\textsuperscript{nd}-order polynomial of 3,000 samples of four features with added white Gaussian noise, and the models were trained on seven features, i.e., three of the features were non-informative. We used the (unnormalized) polynomial map, so that the target function was included in the hypotheses set of the CP-based predictor (for $d\geq 3$). As a preprocessing step, the features were standardized to zero mean and unit variance, and the mean squared error (MSE) was used as the loss function. Given a mini-batch $\mathcal{S}$ of size $S$, the loss function we used for the CP-based predictor is given by
\begin{align}
    L=\frac{1}{S}\sum_{\{\mathbf{x}, y\}\in \mathcal{S}}(f(\mathbf{x})-y)^2+P\left(\mathcal{B}, \mathcal{W}\right),
\end{align}
where $f(\mathbf{x})$ is given in (\ref{prediction}), $P\left(\mathcal{B}, \mathcal{W}\right)$ is given in (\ref{eq:order_reg}), and $y$ are the target labels. 

Moreover, the mini-batch size was set to 32, and the ADAM \cite{adam} optimizer was used. The neural network was composed of three hidden layers, with 20 neurons in the first two and 15 in the last layer, all followed by the ReLU activation and batch normalization \cite{batchnorm}. The test set comprised 20\% of the dataset, while 20\% of the remaining data points formed the validation set. 

\begin{figure}
\centering

    \begin{subfigure}{0.4\textwidth}
       \centering
       \includegraphics[scale=0.5]{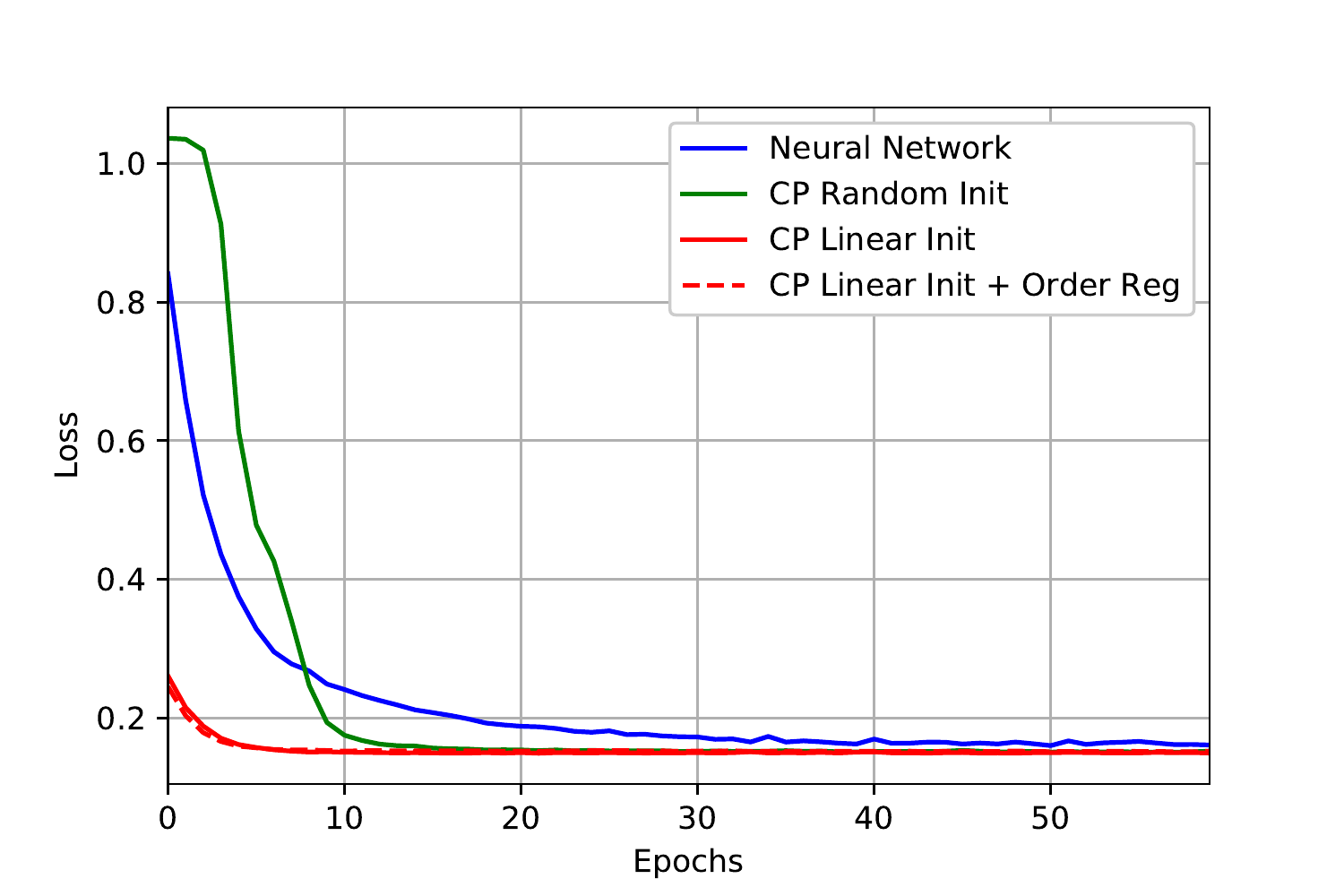}
       \caption{}
    \end{subfigure}          
    \begin{subfigure}{0.4\textwidth}
       \centering
       \includegraphics[scale=0.5]{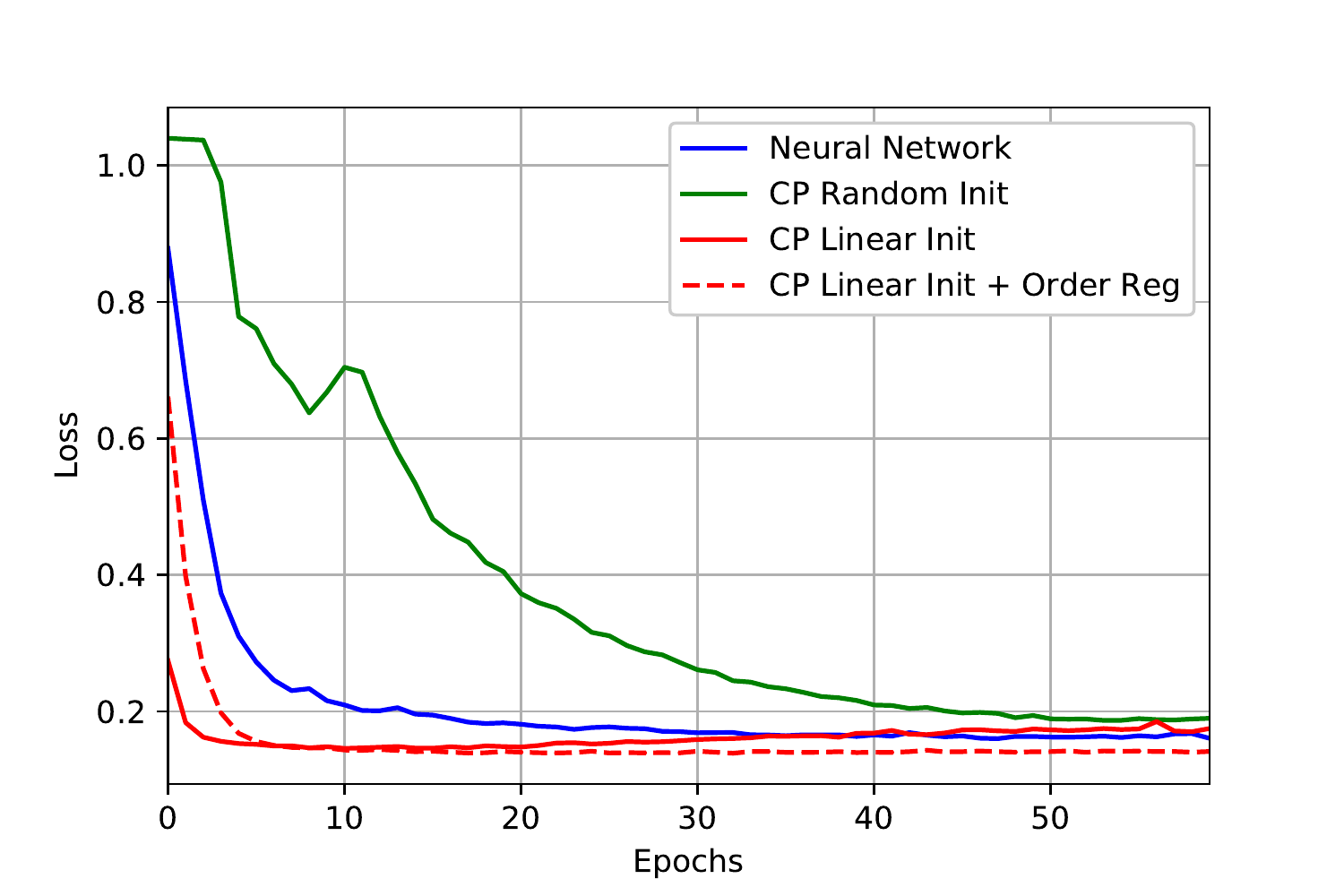}
       \caption{}
       \label{fig:test}
    \end{subfigure}          

\caption{Learning curves of validation MSE for the synthetic polynomial regression task. The effects of initialization with the linear model solution and order regularization are shown for (a) smaller local dimension than the optimal ($d=2$) and $R=7$, and (b) larger local dimension than the optimal ($d=4$) and $R=30$.} \label{fig:art_regression_lc}
\end{figure}

Fig. \ref{fig:art_regression_lc} shows the influence of linear initialization and order regularization on the performance of the proposed model evaluated over a range of local dimensions, with the neural network predictor acting as a baseline. More specifically, we trained the models for 60 epochs, first using random initialization (zero-mean Gaussian noise with standard deviation of 0.2) and no regularization, then with the linear model solution as initialization, and finally order regularization was included with $\beta=3.0$ and $\alpha=10^{-3}$. The top panel of Figure \ref{fig:art_regression_lc} shows the learning curves for $d=2$ and $R=7$, and illustrates that the model was not expressive enough to capture the 2\textsuperscript{nd}-order terms of the polynomial. Despite the linear-initialized model starting from a significantly lower validation error and converging faster, both initialization schemes led to the model reaching the same error after 4-5 epochs; the neural network converged after about 60 epochs to a similar value. Also, order regularization did not have a significant effect on the performance of the model. On the other hand, when $d$ was increased to 4 (a higher than optimal local dimension, as there were no 3\textsuperscript{rd}-order terms) and $R$ to 30, Figure \ref{fig:test} shows that the random initialization scheme took significantly longer to converge (around 60 epochs), and order regularization enabled the optimization procedure to reach a lower validation error.

A comparison with the other considered predictors is given in Table \ref{table:MSE_artif}, which shows the minimum validation score across 100 epochs and training times. We were not able to run the TT-based predictors with the unnormalized polynomial mapping (neither Google's TensorNetwork nor Exponential Machines with $d=2$), as the loss diverged with all hyperparameters we tried.

\begin{table}
\caption{Validation losses and training times for the synthetic regression dataset.}
\centering
{\begin{tabular}{l c c c }\hline
 Method & Val MSE & Train Time (sec)  \\ \hline
CP-Based ($d=3$) (ours)  & \textbf{0.1378} &  11.09  \\
Linear Regression & 0.3311 & \textbf{0.004} \\
RBF SVR & 0.1790 & 1.30 \\
Neural Network & 0.1545 & 20.59 \\
3\textsuperscript{rd}-order Polynomial Network & 0.1390 & 6.96 \\
3\textsuperscript{rd}-order Factorization Machine  & 0.1485 & 4.35 \\
\hline
\end{tabular}}
\label{table:MSE_artif}
\end{table}

\subsection{Performance on Recommender Systems}
Categorical data are challenging for traditional neural networks and SVMs due to data sparsity after one-hot encoding. For example, SVMs with a polynomial kernel find it difficult to learn the coefficients corresponding to interactions of two or more categorical features, because there are usually not enough training points where the relevant binary, one-hot features are both ``hot.'' In contrast, the CP-based predictor factorizes these coefficients and, thus, performs well in such settings (for a detailed discussion on why factorized models perform well with sparse data see \cite{rendle2010factorization}).

We next demonstrate the performance of our model on the MovieLens 100K, a widely-used benchmark dataset for sparse data classification; for more details see \cite{Harper:2015:MDH:2866565.2827872}. For this experiment, we adapted the code from \cite{novikov2018exponential}, in order to be able to directly compare the results with their TT-based predictor. The features were mapped as described in Section \ref{subsec:map_cat}. The CP rank was set to 30, and order regularization was added with $\beta=3.6$ and $\alpha=5\times 10^{-5}$, respectively, while MSE was used as the loss function. Results are shown in Table \ref{rec}. We trained the proposed CP-based model until convergence, obtaining the highest Area Under Curve (AUC) score between all considered models.  For Tensor Train, we report the score stated in \cite{novikov2018exponential}, as well as the score we obtained when running their code locally. For the Tensor Ring (TR) and Hierarchical Tucker (HT) models, we report the scores from our implementations; note that this is the first time that these two Tensor Network formats are considered in this framework. For a comparison with other models, we refer the reader to \cite{novikov2018exponential}. We found that both initialization from the linear model solution and order regularization helped significantly in achieving better performance on this task.

Finally, we also attempted to fit a Tucker-based predictor. However, since in this case we have 26 features, the core tensor in the Tucker format is of order 26, which led to an ``out of memory'' error.

\begin{table} 
\caption{Validation AUC for the MovieLens 100K dataset.}
\centering
{\begin{tabular}{l c c  }\hline
 Method & Val AUC  \\ \hline
Logistic Regression & 0.7821  \\
CP-based (ours)  & \textbf{0.7863}   \\
Exponential Machines \cite{novikov2018exponential} & 0.784  \\
TT-based (local) & 0.7815 \\
TR-based  & 0.7827  \\
HT-based & 0.7803 \\
\hline
\end{tabular}}
\label{rec}
\end{table}

\subsection{Effect of Local Dimension}
\begin{figure}
\centering
       \includegraphics[scale=0.6]{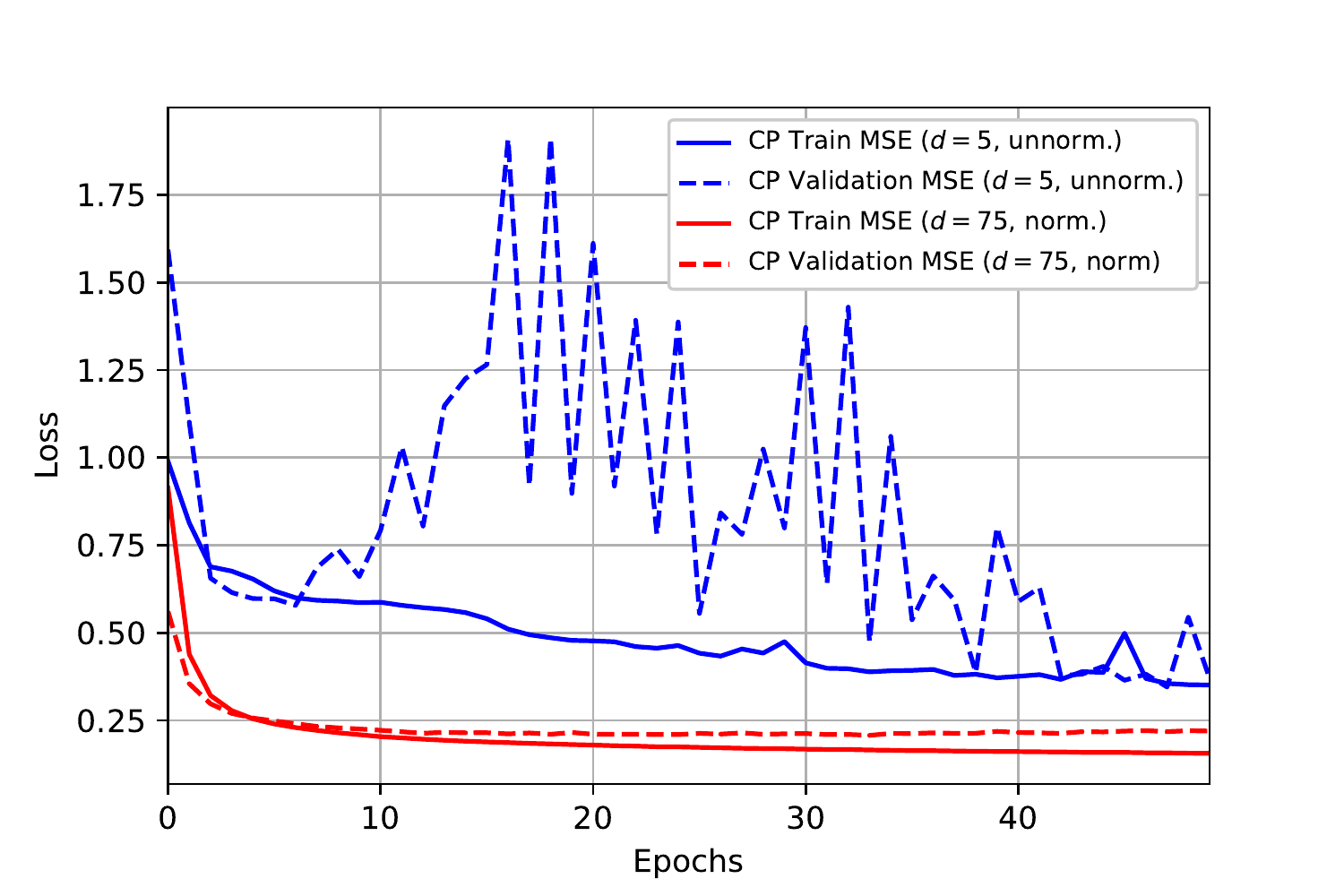}
 \caption{Learning curves for $d=5$ with the unnormalized polynomial map and for $d=75$ with the normalized polynomial map for the California Housing dataset.} \label{fig:curves}
\end{figure}

To show the effect of a high local dimension $d$ on a common single-target regression dataset, we employed the California Housing dataset, comprising 8 features and 20,640 samples, and with similar data pre-processing as for the synthetic example; the CP rank was set to 20. We found that, for the unnormalized polynomial map, an increase in $d$ from 2 to 3 improved performance, but any further increase caused the learning procedure to become highly unstable and eventually to experience numerical instability. Standardizing (zero mean, unit standard deviation) each element of the resulting vector after mapping prevented overflow but led to relatively erratic learning curves for validation (see Fig \ref{fig:curves}). 

\begin{figure}
\centering
       \includegraphics[scale=0.6]{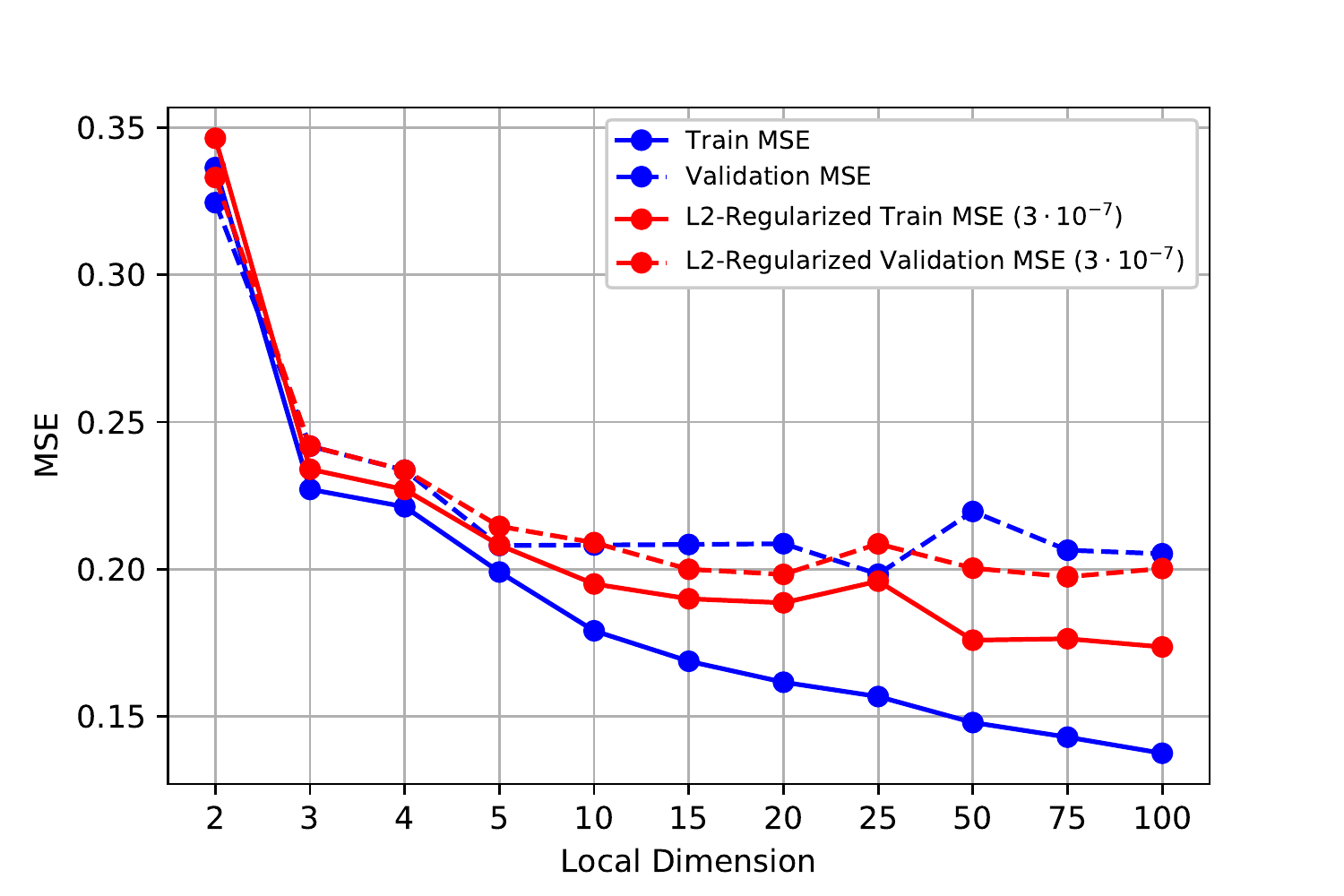}
 \caption{Validation loss for the California Housing dataset as a function of the local dimension, with and without L2 regularization.} \label{fig:high_d}
\end{figure}

In contrast, the model with the normalized polynomial map (with random initialization\footnote{It is also possible to use initialization from the linear model.}) was able to be trained reliably even for very high $d$, as indicated by the smooth learning curves in Fig. \ref{fig:curves}. In addition, without regularization the training error kept decreasing to very small values, thus reflecting the gains in expressiveness (see Fig. \ref{fig:high_d}). The lowest validation error (without regularization) was obtained with $d=25$, and it remained low even for local dimensions around $d=100$, indicating the regularization capabilities of low rank. Of course, with a smaller dataset, we would have observed more overfitting for high $d$. With L2 regularization on the factor matrices, the validation error reached its smallest value at $d=75$ (we did not observe any improvement using order regularization over L2 with this map and on this dataset).

\begin{figure} [h]
\centering
       \includegraphics[scale=0.6]{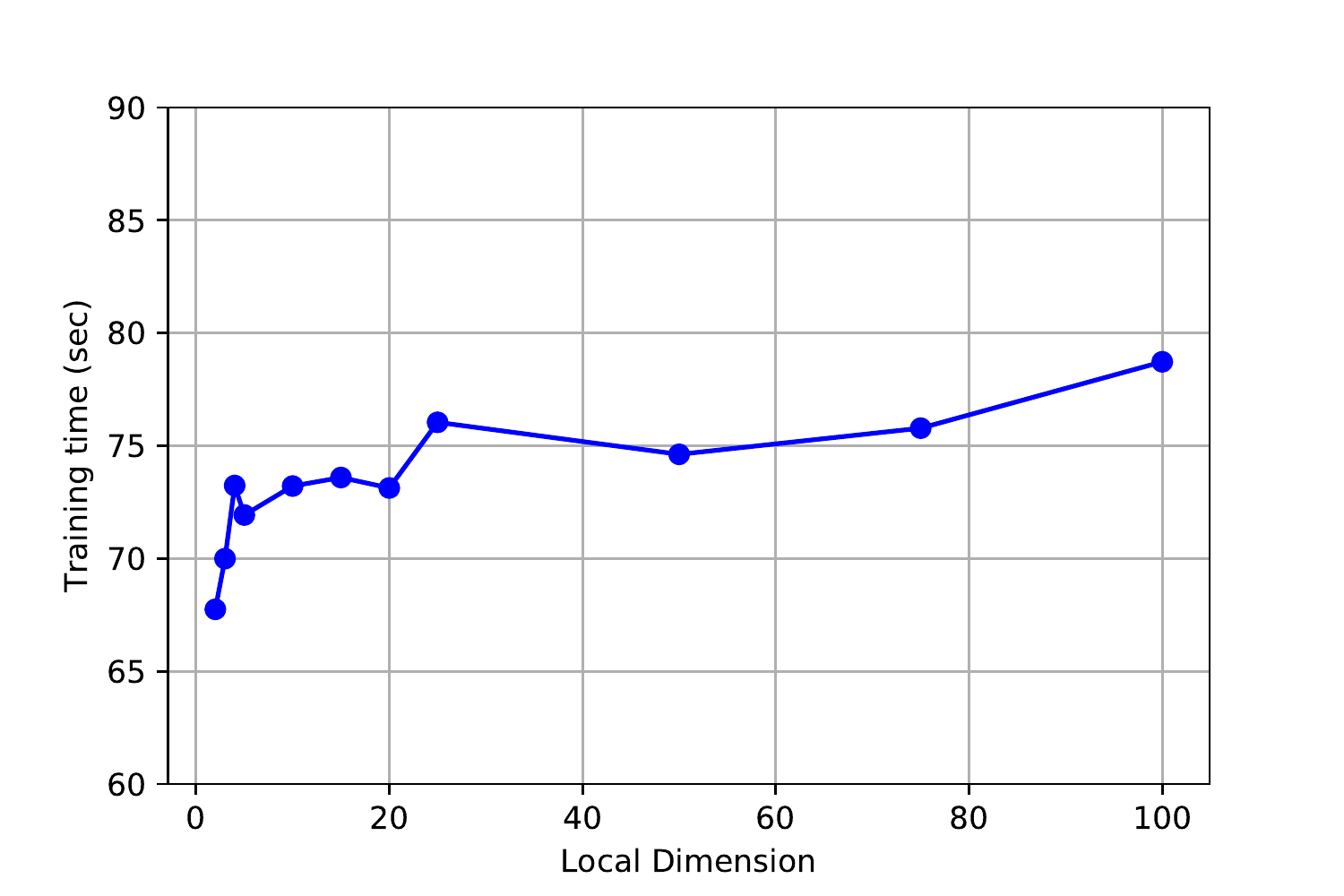}
 \caption{Training time in seconds for the California Housing dataset as a function of the local dimension (run for 50 epochs and CP rank equal to 20).} \label{fig:times}
\end{figure}

A higher $d$ not only resulted in significant gains in performance, but it also had a small impact on training time, due to optimized matrix-vector multiplications in \texttt{TensorFlow}, enabling fast tuning of this hyperparameter (see Fig. \ref{fig:times}).

\begin{table}
\caption{Validation losses and training times for the California Housing dataset.}
\centering
{\begin{tabular}{l c c c }\hline
 Method & Val MSE & Train Time (sec)  \\ \hline
Linear Regression & 0.3712 & \textbf{0.023} \\
CP-based ($d=75$, norm., w/ L2) (ours) & \textbf{0.1959} & 118.84 \\
TT-based ($d=75$, norm., w/ L2) & 0.2128 & 773.25  \\
Tucker-based ($d=75$, norm., w/ L2) & 0.2084 & 659.86  \\
TR-based ($d=75$, norm., w/ L2) & 0.2056 & 249.53 \\
HT-based ($d=75$, norm., w/ L2) & 0.1999 & 113.32 \\

RBF SVR & 0.2236 & 18.78 \\
Neural Network & 0.2029 & 153.20  \\
4th order Polynomial Network & 0.2761 & 9.06 \\
4th order Factorization Machine  & 0.3322 & 51.69 \\
\hline
\end{tabular}}
\label{table:MSE_Calif}
\end{table}

A comparison with the other considered models is given in Table \ref{table:MSE_Calif}. The TT and TR ranks were set to 5, while the HT rank was set to 10, leading to the same number of parameters as in the CP-based predictor. Tucker rank was set to 5. The fully-connected neural network contained four hidden layers and also had an equal number of parameters as the tensor models. Table \ref{table:MSE_Calif} shows the minimum validation score across 100 epochs, and indicates that the CP-based predictor maintained the best performance over all hyperparameters we experimented with. Note that the very large training time of the TT-based predictor compared to the CP-based predictor is likely due to the different implementation in \texttt{TensorFlow}, and not due to the TT-based predictor being inherently slower to train for the same number of parameters. On the other hand, slow training time of the Tucker predictor despite the small number of features reflects its incompatibility with the proposed framework, as elaborated in Section \ref{CP_frame}.

\subsection{Effect of CP Rank}
\begin{figure} 
\centering
       \includegraphics[scale=0.6]{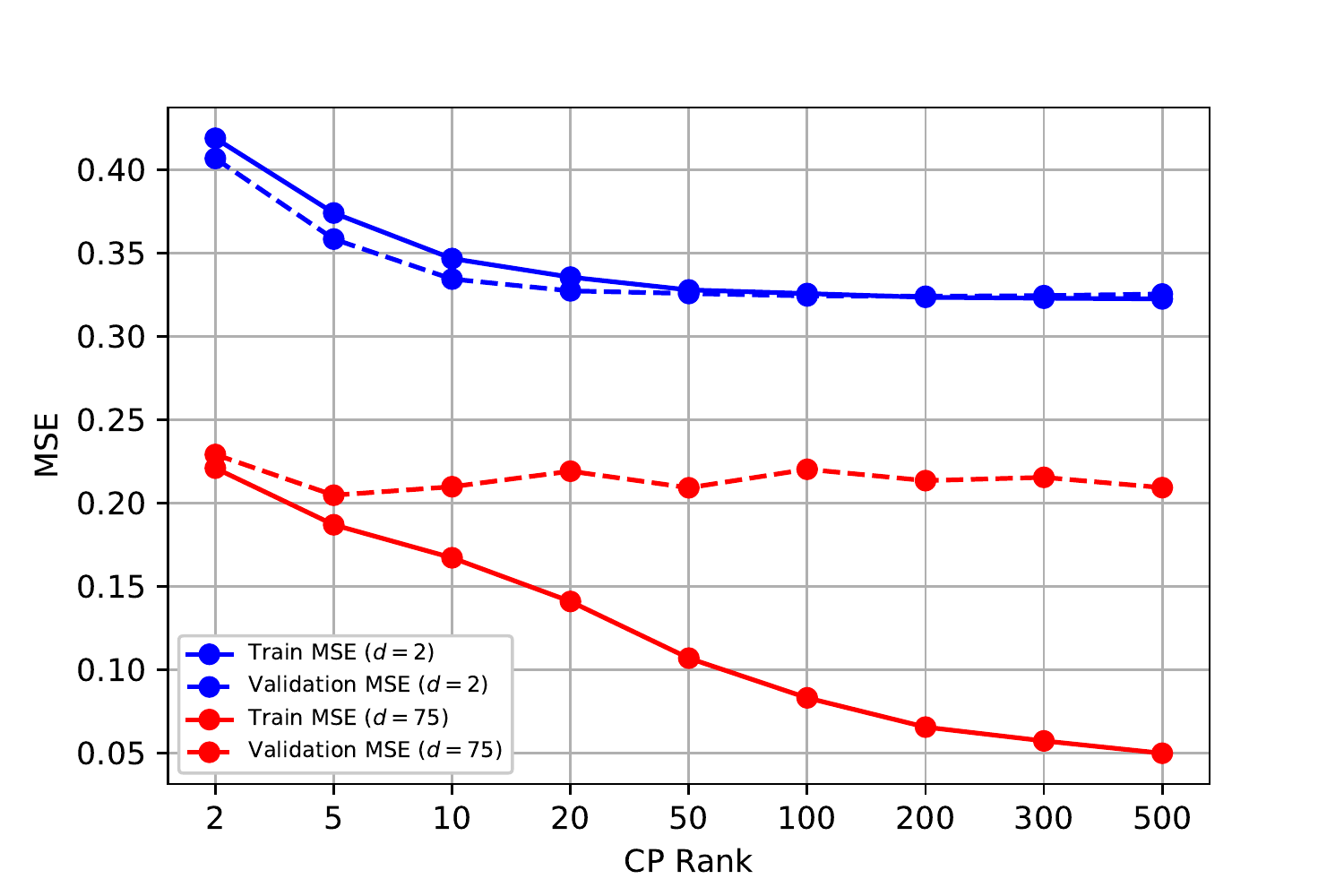}
 \caption{Training and validation losses as functions of the CP rank for low and high local dimensions.} \label{fig:ranks}
\end{figure}

The CP rank affects performance differently for low $d$ than for high $d$. As can be seen in Fig. \ref{fig:ranks}, both the training and validation errors were decreasing up to rank equal to 50 and then plateaued for $d=2$. On the other hand, when $d$ was increased to 75, although the training error was decreasing with higher rank, the validation error remained fairly constant. Notice that even with very small rank (about 2 to 5), the validation error still remained significantly lower than for any value of rank with $d=2$. \textit{This suggests that the local dimension is more important than the rank for accurate model predictions}.

\subsection{Effect of Large Number of Features}
Increasing the number of features beyond a certain point (e.g., about 30) often makes the optimization process more sensitive to the width of the Gaussian distribution when random initialization is used. This stems from the many Hadamard products (or matrix-matrix products in the case of TT-based predictors) that are performed; if the standard deviation of the Gaussian is too small or too big, this can lead to vanishingly small or exceedingly large predictions (and gradients), respectively. Initialization with a linear model solution largely alleviates this issue, as was confirmed when the model was trained on the (flattened) MNIST dataset with 784 features. It would be interesting to investigate other strategies that could enable these models to achieve state-of-the-art results on very high-dimensional datasets, a subject of future work.

\section{Conclusions and Future Work} \label{sec:conclusion}

We have introduced highly efficient and scalable inference and learning algorithms for a non-sequential supervised learning paradigm. This has been achieved based on the Canonical Polyadic Decomposition, which has been shown to serve as a physically meaningful and superior alternative to the existing methods based on the Tensor Train (TT). For rigor, novel predictors based on the Tucker, Tensor Ring, and Hierarchical Tucker formats have also been introduced, with the proposed model outperforming all of the other Tensor Network models on both sparse and dense data. By virtue of multilinear algebra, we have also derived efficient procedures to incorporate order regularization and have established robust model initialization. In addition, a unit-normalized version of an arbitrarily high-dimensional local feature map has been proposed, which enables a straightforward increase in model expressiveness and remains stable for very high dimensions.

Future research directions include extensions based on the Riemannian optimization approach of Exponential Machines, to handle high local dimensions for tensor-based supervised learning, while another promising direction is to explore other local feature maps. Finally, the generalization ability of the models may be enhanced by employing different Tensor Networks to represent the weight tensor, depending on the nature of the data at hand. For example, given the inherent one-dimensional structure of natural language processing problems or the two-dimensional nature of images, it seems likely that the TT representation (1-D TNs) would lead to superior performance in the former case while PEPS \cite{evenbly2011tensor} (2-D TNs) would be more suitable in the latter case.


%

\appendices
\section{Detailed Comparison With Closely-Related Predictors} \label{app:related_works}

In this appendix, we discuss in detail similarities and differences between the proposed CP-based predictor and related predictors.
\subsection{TT-based Models} \label{subsec:ttbased}
The work in \cite{NIPS2016_6211} and \cite{novikov2018exponential} both used the TT format to reduce the number of model parameters. To better understand how the TT-based predictors are linked to the CP-based one, it is important to notice that a CP decomposition can be expressed in terms of the TT format. Specifically, a tensor in the CP format with rank $R$ is equivalent to a tensor in the TT format, for which the TT cores are given by $\mathcal{G}^{(1)}=\mathbf{A}^{(1)}$, $\mathcal{G}^{(N)}=\mathbf{A}^{(N)T}$, and $\mathbf{G}_{i_n}^{(n)}=\diag(a_{i_n,1},\dots,a_{i_n,R})$ for $n=2,\dots,N-1$, where $\mathbf{G}_{i_n}^{(n)}$ are the lateral slices of the cores \cite{phan2019tensor}. Hence, any CP-decomposed weight tensor can be converted into its TT format. However, the predictions produced by the models are not equivalent, since an optimization process on the TT-based predictor alters the off-diagonal terms of the core tensors, i.e., they are not constrained to be zero. Thus, the differences in performance between the two models arise due to off-diagonal core elements being either helpful or detrimental to the generalization capabilities for the dataset at hand. Viewed another way, for an equal number of parameters, the TT-based predictor will achieve superior results if and only if it is better for generalization to have TT cores whose slices are relatively small matrices with off-diagonal terms rather than larger diagonal matrices. Note also that any tensor in the TT format can be mapped to a tensor in the CP format, but, in general, the CP rank equals $R_1R_2\cdots R_{N-1}$, while $\text{TT rank}=\{R_1,\dots,R_{N-1}\}$ \cite{phan2019tensor}. 

A conceptual advantage of the CP-based predictor for non-sequential data is its insensitivity to the ordering of the features, as, when constrained to be laterally diagonal, the cores can be permuted in any way without altering the decomposition (since diagonal matrices commute). On the other hand, the TT-based predictors may be more suitable for sequential data due to the inherent ordering of their cores.

\subsection{Kernel SVM}
The CP-based predictor with a polynomial feature map resembles an SVM with the polynomial kernel, the prediction of which is given by
\begin{align}
    \hat{y}(\mathbf{x})=\left\langle\psi(\mathbf{x}),\mathbf{w}_{\text{svm}}\right\rangle,
\end{align}
where $\psi$ maps the feature vector $\mathbf{x}$ onto a higher-dimensional space, and $\mathbf{w}_{\text{svm}}$ are the parameters of the SVM model. Recall that an SVM with a polynomial kernel, $K(\mathbf{x},\mathbf{z})=\langle\psi(\mathbf{x}),\psi(\mathbf{z})\rangle=\left(\langle\mathbf{x},\mathbf{z}\rangle+1\right)^i$, can model an $i$\textsuperscript{th}-degree polynomial, similarly to our model (with $d=i+1$). However, the disadvantages of the polynomial SVM compared to the CP-based predictor, are:
\begin{itemize}
    \item At least quadratic scaling with the training set size;
    \item Tendency to overfit for large $i$, since, unlike our model, the SVM parameters are independent of one another;
    \item Inability to effectively capture interactions for sparse (categorical) data (e.g., recommender systems);
    \item Compromised physical interpretability since one cannot recover the coefficients of the polynomial;
    \item Predictions which depend on the training data, or support vectors.
\end{itemize}
\subsection{Higher-Order Factorization Machines and Polynomial Networks}
Higher-Order Factorization Machines (HOFM) \cite{rendle2010factorization} address the limitations of the SVMs by factorizing the interaction parameters. In this case, the order refers to the highest degree of feature interactions being modelled (e.g., an order of three refers to modelling interactions containing a maximum of three variables). Although a HOFM resembles the CP-based predictor when $\phi(x_n)=[1,x_n]^T$, there are some important differences, which can be illuminated by casting HOFM into the tensor format \cite{blondel2016polynomial}.

A HOFM of order $L$ can be expressed in the tensor format as
\begin{align}
    \hat{y}=\sum_{l=1}^L\sum_{j_l>\dots>j_1}\hat{w}^{(l)}_{j_1,\dots,j_l}\prod_{k=1}^lx_{j_k}, \label{eq:hofm}
\end{align}
where the weight tensors $\hat{\mathcal{W}}^{(l)}\in\mathbb{R}^{N_l}$ ($l=1,\dots,L$) are represented in the \textit{symmetric} CP format. Furthermore, in this formulation we observe the outer product of the whole feature vector $\mathbf{x}$ with itself $L$ times. Finally, only the entries above the super-diagonal of the weight tensors (which correspond to the products of distinct features) are used to construct the output. In contrast, the proposed CP-based predictor takes the inner product between a tensor assumed to be in the (asymmetric) CP format (and of order equal to the number of features) and a tensor formed from the outer product of the local feature maps $\phi(x_n)=[1,x_n]^T$.

Similarly, Polynomial Networks (PN) \cite{livni2014computational} can be cast into the form \cite{blondel2016polynomial}:
\begin{align}
    \hat{y}=\sum_{l=1}^L\sum_{j_1,\dots,j_l}\hat{w}^{(l)}_{j_1,\dots,j_l}\prod_{k=1}^lx_{j_k},
\end{align}
with the difference from (\ref{eq:hofm}) being in the subscript of the second sum; this is related to our model when $\phi(x)=[1,x,x^2,\dots,x^{(d-1)}]^T$ in the same way as described for HOFM.

Unlike HOFM and PN, the CP-model used in our work allows for the modelling of all-order interactions with a computational complexity that scales linearly with the number of features during both training and inference.

\subsection{Convolutional Arithmetic Circuits}
The CP-based predictor can be viewed as a special case of the model presented in \cite{cohen2016expressive} (see Section 3 in \cite{cohen2016expressive}), where a data sample is represented as a collection of vectors, each of dimensionality $s$. In the case of a grayscale image, these vectors, for example, may correspond to $s$ consecutive pixels. The so-called representation functions (analogous to the local feature maps) are then applied to each of these vectors, using for example an affine transformation followed by an activation function. This model is referred to in \cite{cohen2016expressive} as the Convolutional Arithmetic Circuit (CAC) due to the nature of the representation function. The main differences between the CP-based predictor and CAC are that we use $s=1$ and the representation functions that we apply are those given in Section \ref{sec:feat_mappings}.

It was proved in \cite{cohen2016expressive} that a model based on the Hierarchical Tucker (HT) decomposition is exponentially more expressive than that based on CPD; that is, an HT-based model realizes functions that would almost always require a CP-based predictor with an exponentially large rank to even approximate them. A similar result was proved in \cite{khrulkov2018expressive}, where a model based on CP format was compared with that based on TT. Although this may at first appear as a disadvantage of the CP-based predictor, in reality this is not the case, at least for non-sequential data. The model architecture captures an inductive bias about the task at hand, and due to either potential overfitting or optimization difficulties, a model that is more expressive than necessary is likely to converge to a solution that is inferior to that obtained by a model well-suited to the data. This is confirmed by our experiments on non-sequential data comparing CP- and other Tensor Network-based predictors. We also show that increasing the rank did not significantly improve the validation loss, which provides further empirical evidence that it is unlikely that the target function for these tasks is such that the CP-based predictor would need an exponentially large rank to approximate.

\section{Model Interpretability and Universal Approximation Property}
\subsection{Model Prediction and Interpretability}
The prediction of the proposed model can be expressed in terms of the rows of the factor matrices of the CPD and the corresponding elements of the feature mapping, in the form
\begin{align}
    f(\mathbf{x})&=\sum_{i_1,\dots,i_N}w_{i_1,\dots,i_N}\prod_{k=1}^N\phi(x_k)_{i_k} \\
    &= \sum_{i_1,\dots,i_N}\left(\bigcircledast_{k=1}^N\hat{\mathbf{a}}^{(k)}_{i_k}\right)\mathbf{1}\prod_{k=1}^N\phi(x_k)_{i_k}. \label{eq:mod_pred}
\end{align}
In other words, the coefficient for each (transformed) feature interaction can be readily obtained after training by simply performing a Hadamard product of the corresponding row vectors of the learned factor matrices and then by summing over all entries of the resulting vector. This provides enhanced interpretability over SVMs or deep learning techniques, together with efficiency, as the importance of a given feature interaction can be observed in $\mathcal{O}(NRd)$ time. 
\begin{example}
Consider again $\phi(x_n)=\begin{bmatrix}
1 , x_n\end{bmatrix}^T$ with $N=3$. Then, the coefficient for e.g., $x_1x_3$ could be obtained by performing the Hadamard product of $\hat{\mathbf{a}}^{(1)}_2$, $\hat{\mathbf{a}}^{(2)}_1$, and $\hat{\mathbf{a}}^{(3)}_2$, and then summing over the entries.
\end{example}
\subsection{Universal Approximation Property}
Recall that the TN framework discussed in Section \ref{TN_framework} can approximate any function with an arbitrary approximation error, as long as the local feature dimension $d$ is large enough \cite{cohen2016expressive}. Since there always exists a CP rank for which a tensor can be exactly decomposed in its CP form, the proposed model also satisfies the universal function approximation property. This highlights the importance of using a higher local dimension $d$ for the feature map, rather than constraining it to the original $d=2$.

\section{Proof of the Khatri-Rao Property} \label{Khatri}
To prove the identity in (\ref{KhatriProp}), let $\mathbf{A}^{(k)}\in\mathbb{R}^{I_k\times J}$ and $\mathbf{B}^{(k)}\in\mathbb{R}^{I_k\times L}$, $1\leq k\leq N$. Since
\begin{align}
    \left(\bigodot_{k=1}^N \mathbf{A}^{(k)} \right)^T=\begin{bmatrix}  \bigotimes_{k=1}^N \mathbf{a}_1^{(k)} && \cdots && \bigotimes_{k=1}^N \mathbf{a}^{(k)}_J \end{bmatrix}^T \nonumber,
\end{align}
\begin{align}
    \left(\bigodot_{k=1}^N \mathbf{B}^{(k)} \right)=\begin{bmatrix}  \bigotimes_{k=1}^N \mathbf{b}_1^{(k)} && \cdots && \bigotimes_{k=1}^N \mathbf{b}^{(k)}_L \end{bmatrix} \nonumber,
\end{align}
and using the mixed-product property \cite{zhang2013}
\begin{align}
    \left(\bigotimes_{k=1}^N \mathbf{a}_j^{(k)T}\right)\left(\bigotimes_{k=1}^N \mathbf{b}_j^{(k)}\right)=\bigotimes_{k=1}^N \mathbf{a}_j^{(k)T}\mathbf{b}_j^{(k)}
\end{align}
we obtain
\begin{align}
    &\left(\bigodot_{k=1}^N \mathbf{A}^{(k)} \right)^T\left(\bigodot_{k=1}^N \mathbf{B}^{(k)} \right) \nonumber \\
    &=\begin{bmatrix} \bigotimes_{k=1}^N \mathbf{a}_1^{(k)T}\mathbf{b}_1^{(k)} && \cdots && \bigotimes_{k=1}^N \mathbf{a}_1^{(k)T}\mathbf{b}_L^{(k)} \\ 
    \vdots && \ddots && \vdots \\
    \bigotimes_{k=1}^N \mathbf{a}_J^{(k)T}\mathbf{b}_1^{(k)} && \cdots && \bigotimes_{k=1}^N \mathbf{a}_J^{(k)T}\mathbf{b}_L^{(k)} \nonumber
    \end{bmatrix} \\
    &=\begin{bmatrix} \prod_{k=1}^N \mathbf{a}_1^{(k)T}\mathbf{b}_1^{(k)} && \cdots && \prod_{k=1}^N \mathbf{a}_1^{(k)T}\mathbf{b}_L^{(k)} \\ 
    \vdots && \ddots && \vdots \\
    \prod_{k=1}^N \mathbf{a}_J^{(k)T}\mathbf{b}_1^{(k)} && \cdots && \prod_{k=1}^N \mathbf{a}_J^{(k)T}\mathbf{b}_L^{(k)} \nonumber
    \end{bmatrix} \\
    &=\bigcircledast_{k=1}^N\mathbf{A}^{(k)T}\mathbf{B}^{(k)}.
\end{align}
\section{Proof of Matrix Calculus Identity}
To prove the identity in (\ref{eq:mat_calc_identity}), let $\mathbf{Y}=\mathbf{X}\circledast\mathbf{W}$ and a colon ($:$) denote the inner product operator. Then,
\begin{align}
    q&=\Tr\left(\mathbf{Y}\mathbf{Z}\mathbf{Y}^T\right) \nonumber \\
     &=\Tr\left(\mathbf{Y}^T\mathbf{Y}\mathbf{Z}\right) \nonumber \\
     &=\mathbf{Z}:\mathbf{Y}^T\mathbf{Y}.
\end{align}
We can now obtain the differential
\begin{align}
    dq&=\mathbf{Z}:d(\mathbf{Y}^T\mathbf{Y}) \nonumber \\
      &=\mathbf{Z}:\left(\mathbf{Y}^Td\mathbf{Y}+d\mathbf{Y}^T\mathbf{Y}\right) \nonumber \\
      &=\mathbf{Z}:\mathbf{Y}^Td\mathbf{Y}+\mathbf{Z}:d\mathbf{Y}^T\mathbf{Y} \nonumber \\
      &=\mathbf{Z}:\mathbf{Y}^Td\mathbf{Y}+\mathbf{Z}^T:\mathbf{Y}^Td\mathbf{Y} \nonumber \\
      &=\left(\mathbf{Z}+\mathbf{Z}^T\right):\mathbf{Y}^Td\mathbf{Y} \nonumber \\
      &=\mathbf{Y}\left(\mathbf{Z}+\mathbf{Z}^T\right):d\mathbf{Y} \nonumber \\
      &=\mathbf{Y}\left(\mathbf{Z}+\mathbf{Z}^T\right):\mathbf{W}\circledast d\mathbf{X} \nonumber \\
      &=\left(\mathbf{Y}\left(\mathbf{Z}+\mathbf{Z}^T\right)\right)\circledast\mathbf{W} : d\mathbf{X}.
\end{align}
Finally, this implies that
\begin{align}
    \frac{\partial q}{\partial \mathbf{X}}=\left(\Big(\mathbf{X}\circledast\mathbf{W}\Big)\Big(\mathbf{Z}+\mathbf{Z}^T\Big)\right)\circledast\mathbf{W}. 
\end{align} \label{Proof}

\section*{Acknowledgments}

We wish to thank the anonymous reviewers for their constructive and insightful comments. We would also like to thank Giuseppe Calvi and Bruno Scalzo Dees for helpful discussions. A.H. is supported by
an Imperial College London President’s Scholarship. K.K. is
supported by an EPSRC International Doctoral Scholarship.

\ifCLASSOPTIONcaptionsoff
  \newpage
\fi



\bibliographystyle{IEEEtran}
\bibliography{references}
%

%




\end{document}